\documentclass[lettersize,journal]{IEEEtran}

\usepackage{amsmath,amsfonts}
\usepackage{array}
\usepackage{pifont}
\usepackage{amsmath} % amsmath 是数学公式的基础包
\usepackage{bm}      % 引入 \bm 命令
\usepackage[caption=false,font=normalsize,labelfont=sf,textfont=sf]{subfig}
\usepackage{textcomp}
\usepackage{stfloats}
\usepackage{url}
\usepackage{graphicx}
\usepackage{cite}
\usepackage{multicol}
\usepackage{multirow}
\usepackage{booktabs}
\usepackage{setspace}
\usepackage[ruled, linesnumbered]{algorithm2e}
\usepackage{makecell} % 导言区
\usepackage{xcolor}
\usepackage{amsmath}
\DeclareMathOperator{\KL}{KL}
\DeclareMathOperator{\Softmax}{Softmax}
\usepackage{fontawesome5}
\usepackage{tabularx}

\usepackage{colortbl}% 表格颜色支持
\definecolor{Ocean}{RGB}{129,154,254}%定义海蓝色
\definecolor{orange}{RGB}{254,144,95}%定义海蓝色
\definecolor{low}{HTML}{D6EAF8}
\definecolor{high}{HTML}{FADBD8}

\definecolor{green}{HTML}{000000} % 32CD32
\definecolor{yellow}{HTML}{000000}
\usepackage[colorlinks,linkcolor=red, urlcolor=magenta]{hyperref}
\IEEEoverridecommandlockouts
\usepackage{algpseudocode}

\usepackage{arydshln} % for dashed lines

\begin{document}

%%  paper title
% \title{Efficient Cross-View Geo-Localization via Hierarchical Distillation and Information-Theoretic Multi-view Selection Refinement}

% \title{MobileGeo: Resource efficient Drone Cross-View Geo-Localization via Hierarchical Knowledge Distillation and Multi-view Selection Refinement}

\title{MobileGeo: Exploring Hierarchical Knowledge Distillation for Resource-Efficient Cross-view Drone Geo-Localization}

% Authors
\author{Jian Sun, Kangdao Liu, Chi Zhang, Chuangquan Chen,~\IEEEmembership{Member,~IEEE}, Junge Shen, C. L. Philip Chen,~\IEEEmembership{Life Fellow,~IEEE}, and Chi-Man VONG,~\IEEEmembership{Senior Member,~IEEE}

%% left Corner
\thanks{Manuscript received October 20, 2025; This work was supported in part by Shenzhen Science and Technology Innovation Committee under Project SGDX20220530111001006, in part by the Science and Development Fund, Macau under 0118/2024/RIA2 and 0216/2024/AGJ, and in part by the University of Macau under Grant MYRG-GRG2023-00061-FST-UMDF. \textit{(Corresponding authors: Chi-Man VONG; Junge Shen.) } 

Jian Sun, Kangdao Liu and Chi-Man VONG are with the Department of Computer and Information Science, University of Macau, Macau 999078, China. (e-mail: sun.j.ac@connect.um.edu.mo; kangdaoliu@gmail.com; cmvong@um.edu.mo;) 

C. L. Philip Chen is with the School of Computer Science and Engineering, South China University of Technology and Pazhou Lab, Guangzhou 510641, 510335, China. (e-mail: philip.chen@ieee.org)

Chuangquan Chen is with the with the Faculty of Intelligent Manufacturing, Wuyi University, Jiangmen 529020, China (e-mail: chenchuangquan87@163.com).

Chi Zhang and Shen Junge are with the Unmanned System Research Institute, Northwestern Polytechnical University, Xi’an 710072, China (e-mail: tonyz001@163.com; shenjunge@nwpu.edu.cn).

}}

% The paper headers
\markboth{Journal of \LaTeX\ Class Files,~Vol.~14, No.~8, June~2025}%
{Shell \MakeLowercase{\textit{et al.}}: A Sample Article Using IEEEtran.cls for IEEE Journals}

\IEEEpubid{\begin{minipage}{\textwidth}\ \centering
		Copyright \copyright 2024 IEEE. Personal use of this material is permitted. \\
		However, permission to use this material for any other purposes must be obtained 
		from the IEEE by sending an email to pubs-permissions@ieee.org.
\end{minipage}}

\maketitle

\begin{abstract}

Cross-view geo-localization (CVGL) plays a vital role in drone-based multimedia applications, enabling precise localization by matching drone-captured aerial images against geo-tagged satellite databases in GNSS-denied environments. However, existing methods rely on resource-intensive feature alignment and multi-branch architectures, incurring high inference costs that limit their deployment on edge devices. We propose MobileGeo, a mobile-friendly framework designed for efficient on-device CVGL: 1) During training, a Hierarchical Distillation (HD-CVGL) paradigm, coupled with Uncertainty-Aware Prediction Alignment (UAPA), distills essential information into a compact model without incurring inference overhead. 2) During inference, an efficient Multi-view Selection Refinement Module (MSRM) leverages mutual information to filter redundant views and reduce computational load. Extensive experiments demonstrate that MobileGeo outperforms previous state-of-the-art methods, achieving a 4.19\% improvement in AP on University-1652 dataset while being over 5$\times$ more efficient in FLOPs and 3$\times$ faster. Crucially, MobileGeo runs at 251.5 FPS on an NVIDIA AGX Orin edge device, demonstrating its practical viability for real-time on-device drone geo-localization. The code is available at \href{https://github.com/SkyEyeLoc/MobileGeo}{\texttt{https://github.com/SkyEyeLoc/MobileGeo}}.

\end{abstract}

\begin{IEEEkeywords}
Cross-view, Distillation, Mutual Information
\end{IEEEkeywords}

% FIG 1
\begin{figure}
  \centering
  \includegraphics[width=0.9\linewidth]{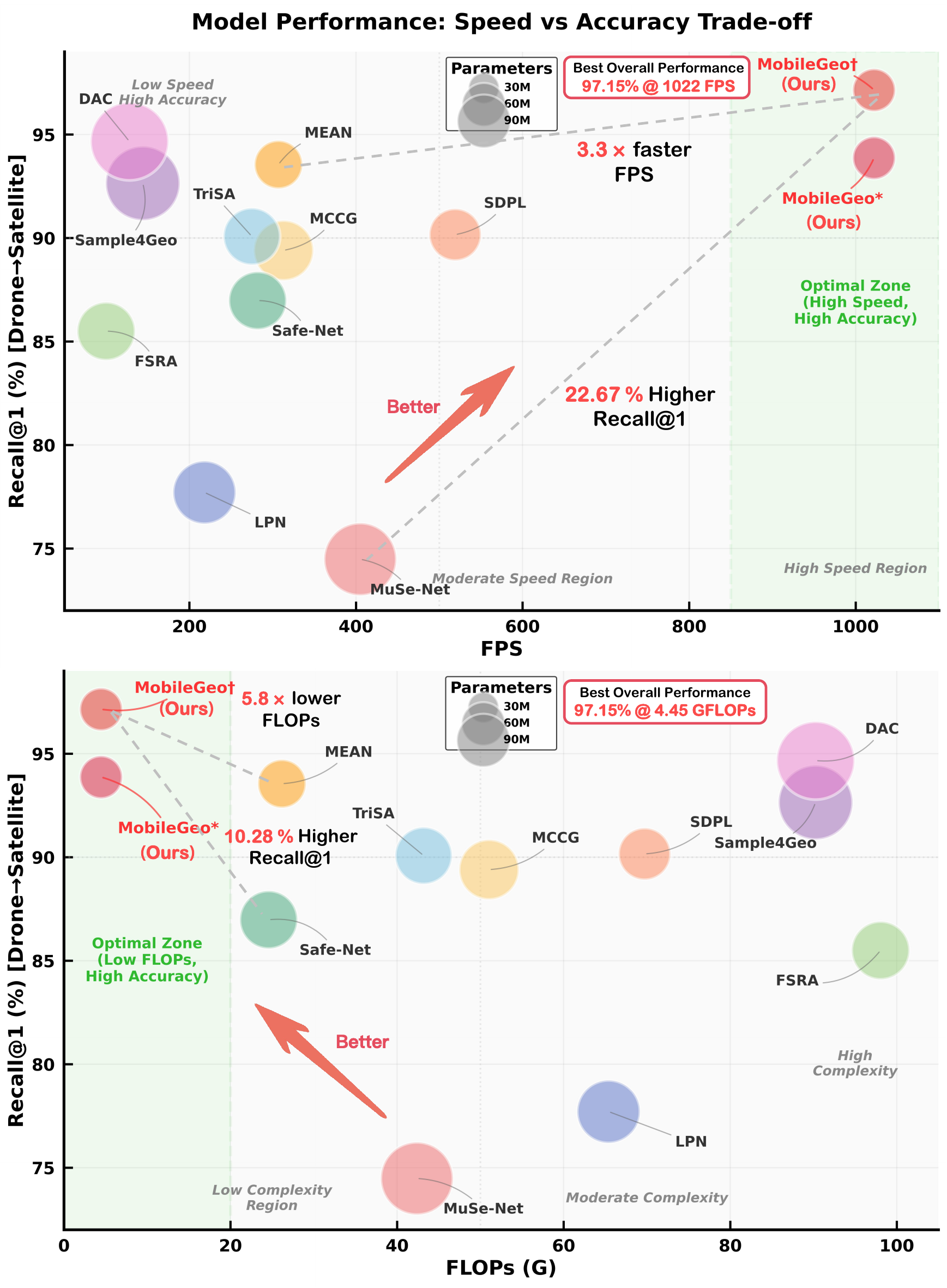}
  \caption{\textbf{Dual-perspective efficiency analysis of our MobileGeo on University-1652 Drone$\rightarrow$Satellite benchmark.} With only 4.45G FLOPs, our approach surpasses heavier models ($>$20G FLOPs) in accuracy. Our method consistently dominates existing approaches in both computational and runtime efficiency while achieving state-of-the-art performance. $*$ denotes the efficient model after hierarchical distillation,  $^\dagger$ indicates the model with post-process.
}
  \label{bubble_all}
  \vspace{-15pt}
\end{figure}

\section{Introduction}
\IEEEpubidadjcol
\IEEEPARstart{}{} 

\IEEEPARstart{C}{ross-view}  geo-localization (CVGL) aims to determine the geographic location of a query image by matching it against a geo-tagged reference database. For drones, this capability is especially critical, offering a pathway to autonomous localization where GPS signals are unavailable. The task typically involves matching multi-view drone images to a corresponding satellite image,  \IEEEpubidadjcol a process complicated by extreme variations in viewpoint and cross-domain appearance.

To improve the cross-view matching precision, the field has rapidly evolved from early methods using handcrafted descriptors to dominant deep learning paradigms built on Siamese or Triplet networks~\cite{zheng2020university,wang2021each,sun2024tirsa,chen2025multi,wu2025ccigeo,zeng2022geo}. More recently, Vision Transformers (ViTs)~\cite{vaswani2017attention} and their variants, such as TransGeo~\cite{zhu2021transgeo} and FSRA~\cite{dai2021transformer}, have set high performance standards by leveraging self-attention to learn powerful, view-invariant global representations. However, despite these advancements, several critical challenges remain that hinder the deployment of these models in practical, real-world mobile scenarios.

% FIG 2
\begin{figure}
  \centering
  \includegraphics[width=0.9\linewidth]{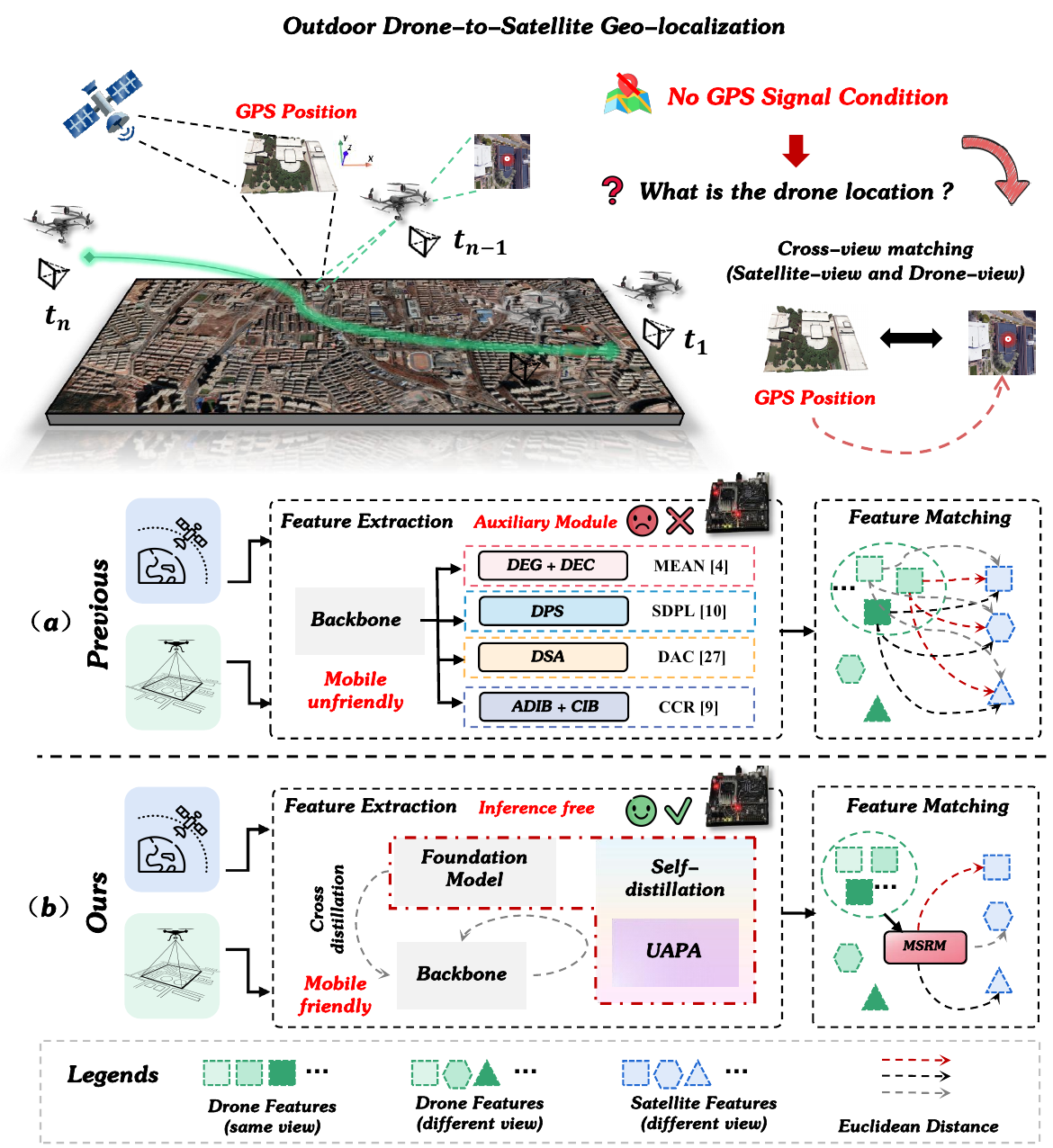}
  \caption{The top panel illustrates the workflow for \textbf{cross-view drone geo-localization in a GPS-signal-denied environment}. The bottom panel contrasts existing approaches with our proposed \textbf{mobile-friendly MobileGeo} method. (a) Illustration of prior methods~\cite{chen2025multi,chen2024sdpl,xia2024enhancing,du2024ccr} that introduce auxiliary modules during feature extraction. (b) Our proposed module is inference-free, incurring no additional computational overhead at deployment. Furthermore, in the feature matching stage, our MSRM significantly reduces computational complexity by selectively filtering and fusing multi-view features. 
}
  \label{compare}
  \vspace{-15pt}
\end{figure}

\textbf{Firstly}, the pursuit of higher accuracy has led to an escalating computational and resource burden. In Figure~\ref{compare}, many state-of-the-art methods achieve superior performance by incorporating complex auxiliary branches, resource-demanding cross-view alignment strategies. \IEEEpubidadjcol As shown in Figure~\ref{bubble_all}, while models like CCR~\cite{du2024ccr} and Mean~\cite{chen2025multi} achieve high recall, they do so with immense computational overhead (e.g., over 90G FLOPs), creating a significant gap between algorithmic advancements and their practical applicability.

\textbf{Secondly}, existing methods~\cite{chen2024sdpl,sun2025cgsi} often lack an explicit mechanism to address the inherent trade-off between semantic abstraction and spatial fidelity. As features pass through deeper layers, they gain semantic robustness at the cost of losing the spatial details (e.g., rooftop textures, landmark patterns) that are critical for discriminating between visually similar locations. Furthermore, the significant data imbalance and domain discrepancy between the different views lead to asymmetric convergence, resulting in suboptimal feature alignment.

\textbf{Thirdly}, current approaches often make inefficient use of multi-view information. While a sequence of drone images provides a rich representation of a landmark, most methods either process each view independently~\cite{zhu2023sues,deuser2023sample4geo} or resort to computationally prohibitive techniques like 3D reconstruction to fuse views~\cite{ju2024video2bev}. The former fails to leverage the collaborative potential of multiple perspectives, while the latter imposes a heavy computational barrier. There is a clear need for a lightweight mechanism that can intelligently select and aggregate the most informative views without significant processing overhead for real-time onboard applications.

Based on the above analysis, a straightforward idea is to simply deploy a lightweight network, but this approach typically leads to a significant performance collapse. To bridge this performance gap, we propose the MobileGeo framework. Overall, this paper makes the following contributions:

\textbf{(1)} We introduce MobileGeo, a novel mobile-friendly method achieves accuracy-efficiency balance in CVGL by concentrating model complexity during training, yielding a highly accurate inference model for resource-constrained devices.

% We propose the Precision-Focused Efficient Design (MobileGeo) paradigm, which resolves the accuracy-efficiency dilemma in CVGL by concentrating model complexity in the training phase, resulting in a highly accurate yet computationally lightweight inference model.

\textbf{(2)} We introduce Hierarchical Distillation for CVGL (HD-CVGL), a novel training framework that synergistically combines inverse self-distillation, uncertainty-aware alignment, and cross-distillation to create a compact feature extractor that excels at capturing both semantic and spatial information.

\textbf{(3)} We propose the Multi-view Selection Refinement Module (MSRM) and provide a theoretical demonstration grounded in mutual information explaining how it enhances localization by optimally selecting and fusing multi-view information while minimizing feature matching overhead.

\textbf{(4)} We conduct extensive empirical evaluations on widely used benchmarks, including University-1652 and SUES-200, demonstrating that MobileGeo establishes a new state-of-the-art in both accuracy and efficiency. Additional deployment on edge devices further validate its real-time capabilities.

\section{Related work}

\subsection{Cross-view drone Geo-localization}

% Cross-view geo-localization involves Siamese or Triplet networks~\cite{lin2015learning, vo2016localizing} to learn a shared, view-invariant embedding space. To explicitly address geometric inconsistencies, some works integrate geometric priors, notably using a polar transform to align satellite and ground-level perspectives~\cite{shi2019where, shi2022optimal} before feature comparison. While ground-to-satellite matching~\cite{wang2023dehi,li2023patch} sets the foundation, the unique oblique perspective of drone introduces a more severe and spatially non-uniform multi-view set of geometric challenges.

Drone-to-Satellite Geo-Localization. This task is particularly challenging due to the multi-view oblique and low-altitude perspective of drone, which creates significant domain gaps between platforms. Following the establishment of key benchmarks~\cite{zheng2020university, zhu2023sues}, research has progressed from improving feature robustness with attention mechanisms~\cite{lai2020understanding,liu2020focus} to the now-dominant Vision Transformer (ViT) architectures. ViTs like TransGeo~\cite{zhu2021transgeo} and FSRA~\cite{cai2022feature} leverage self-attention to learn powerful global representations, setting high performance standards. More recently, the field has explored more efficient and powerful backbone architectures. For instance, several works have successfully employed ConvNeXt~\cite{liu2022convnet} to extract highly discriminative global features, achieving improved performance and efficiency~\cite{guan2024multi}. However, most of contemporary methods~\cite{chen2025multi,xia2024enhancing,11297780} achieve higher accuracy by incorporating sophisticated auxiliary branches or modules, this paradigm is ill-suited for resource-constrained platforms. 

% We follow a precision-focused efficient design paradigm to empower a simple backbone. By strengthening its feature extraction capabilities without adding any inference-time modules, our model achieves competitive precision while drastically reducing the computational burden, making it ideal for deployment on platforms like drone.

% % frame_work
\begin{figure*}
  \centering
  \includegraphics[width=0.9\linewidth]{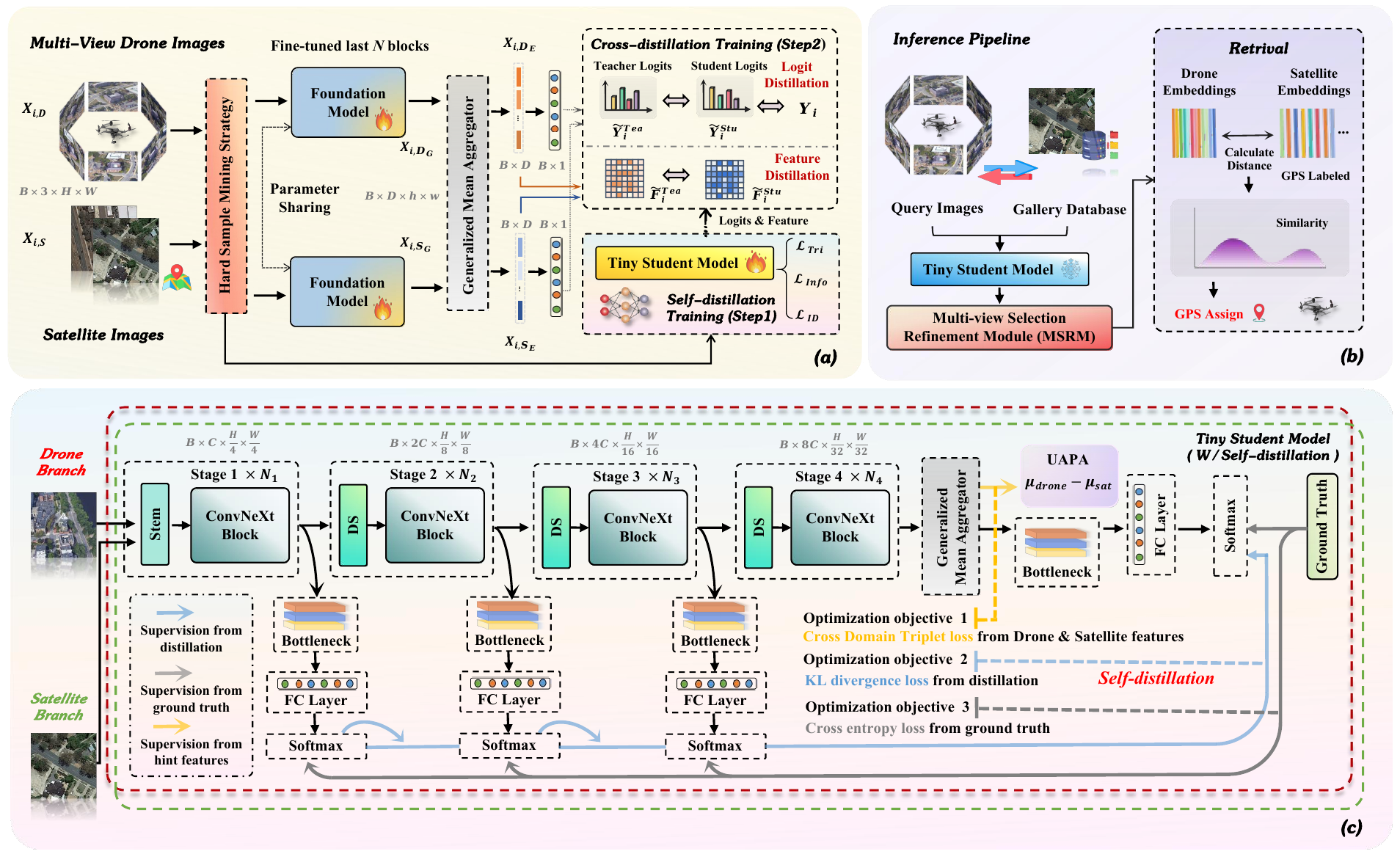}
  \caption{\textbf{Overview of our MobileGeo framework.} (a) The Hierarchical Distillation for CVGL (HD-CVGL). This is a two-step process: first, a tiny student model undergoes Fine-Grained Inverse Self-distillation. Second, the fine-tuned foundation model acts as a teacher, providing guidance at both the feature and logit levels. (b) The inference stage pipeline. The Multi-view Selection Refinement Module (MSRM) leverages mutual information to select discriminative drone images from multiple views, effectively boosting both retrieval accuracy and speed. (c) A detailed illustration of the self-distillation pipeline, depicting the flow of optimization objectives. It incorporates an Uncertainty-Aware Prediction Alignment (UAPA) mechanism to mitigate challenges from data imbalance.}
  \label{frame_work}
  \vspace{-5pt}
\end{figure*}

\subsection{Multi-view Refinement}

Traditional drone-based geo-localization relies on direct matching between individual query images and a reference database~\cite{shen2023mccg,sun2024tirsa,dai2021transformer}. However, these approaches struggle with significant viewpoint discrepancies caused by factors such as occlusions from structures or vegetation, and the diverse perspectives captured by drone operating at varying altitudes and angles, making accurate location recognition increasingly challenging. Recent works have recognized the importance of leveraging multiple drone views to improve localization accuracy with approaches using 3D reconstruction~\cite{li2025unsupervised,zhou2024nerfect,moreau2022lens} to represent scenes from multi-view observations and iteratively refining camera poses to align rendered views with satellite imagery.  This multi-view fusion paradigm~\cite{zhang2021reference} demonstrates significant improvements by exploiting the rich geometric information contained across different viewpoints. 

Mutual information (MI) has emerged as a principled criterion for selecting informative views in a wide variety of complex systems. In 3D reconstruction, MI has guided next-best-view selection effectively~\cite{xie2025gauss}. Similarly, in multi-view clustering, recent work~\cite{zhang2023mutual} minimizes MI between common and view-specific representations to exploit inter-view complementary information to preserve principal information. 

Existing methods process all available views through expensive reconstruction-based methods~\cite{ju2024video2bev,li2025unsupervised}. In contrast to these complex approaches, we propose an efficient MSRM that operates as a lightweight post-processing method. Rather than constructing expensive 3D representations, our method directly aggregates selected features from multiple drone viewpoints through Mutual Information theory.

\section{Proposed Method}

% In this section, we provide a comprehensive introduction to our proposed MobileGeo. As illustrated in Figure \ref{frame_work}, we follow a precision-focused efficient design in two stages: training and inference. During the training stage, we propose Hierarchical Distillation for CVGL (HD-CVGL). Initially, we employ Fine-Grained Inverse Self-distillation (FISD) based on the internal stages of an efficient student model. Concurrently, to address the challenge of data imbalance, we introduce Uncertainty-Aware Prediction Alignment (UAPA), achieving a more robust and stable train. Subsequently, guided by a fine-tuned foundation model, we perform cross-distillation to obtain a student model with higher precision. In the inference stage, we propose a Multi-view Selection Refinement Module (MSRM) to better leverage multi-view information and further reduce the time cost of feature matching.

\subsection{Hierarchical Distillation for CVGL}
\paragraph{\textbf{Fine-Grained Inverse Self-distillation}} In the CVGL task, a fundamental trade-off in designing deep networks exists between semantic abstraction and spatial fidelity. As shown in Figure \ref{frame_work} (c), the student network $\mathcal{N}$, composed of $N$ hierarchical stages ($N=4$), transforms an input image $\mathbf{I}$ into a sequence of feature representations $\{\mathbf{F}_1, \dots, \mathbf{F}_N\}$, where $\mathbf{F}_i \in \mathbb{R}^{C_i \times H_i \times W_i}$. As the depth $i$ increases, $\mathbf{F}_i$ gains semantic abstraction at the cost of losing the fine-grained spatial details present in the shallower features. In cross-view matching, these discarded low-level details often contain critical view-invariant cues (e.g., rooftop textures, landmark patterns).

To ensure the final representation $\mathbf{F}_N$ is both semantically robust and perceptually detailed, we propose a novel method named fine-grained inverse self-distillation (FISD), a form of hierarchical consistency regularization. This approach inverts the conventional knowledge transfer paradigm~\cite{zhang2019your}. Instead of the deep layer teaching the shallow, we compel the final student layer to retain the discriminative knowledge discovered by the shallower "teacher" layers. This inverse knowledge transfer is motivated by two fundamental observations:

\begin{itemize}
    \item \textbf{Targeting the Task-Specific Layer:} It is the final layer's feature that is ultimately used for the matching task. Consequently, this is the representation that must be refined and enriched to maximize performance.

    \item \textbf{Leveraging a Spatial-Detail-Preserving Teacher:} Shallower layers serve as an authoritative teacher by preserving the fine-grained spatial information that deeper, more semantic layers progressively lose to abstraction.
\end{itemize}

We attach an auxiliary classification head $\mathcal{C}_i$ to each stage's feature map $\mathbf{F}_i$, producing logits $z_i = \mathcal{C}_i(\mathbf{F}_i)$. The core of FISD is to align the probability distribution of the final stage, $z_N$, with those from all preceding stages $\{z_1, \dots, z_{N-1}\}$.

% % accuracy_gap
\begin{figure}
  \centering
  \includegraphics[width=0.9\linewidth]{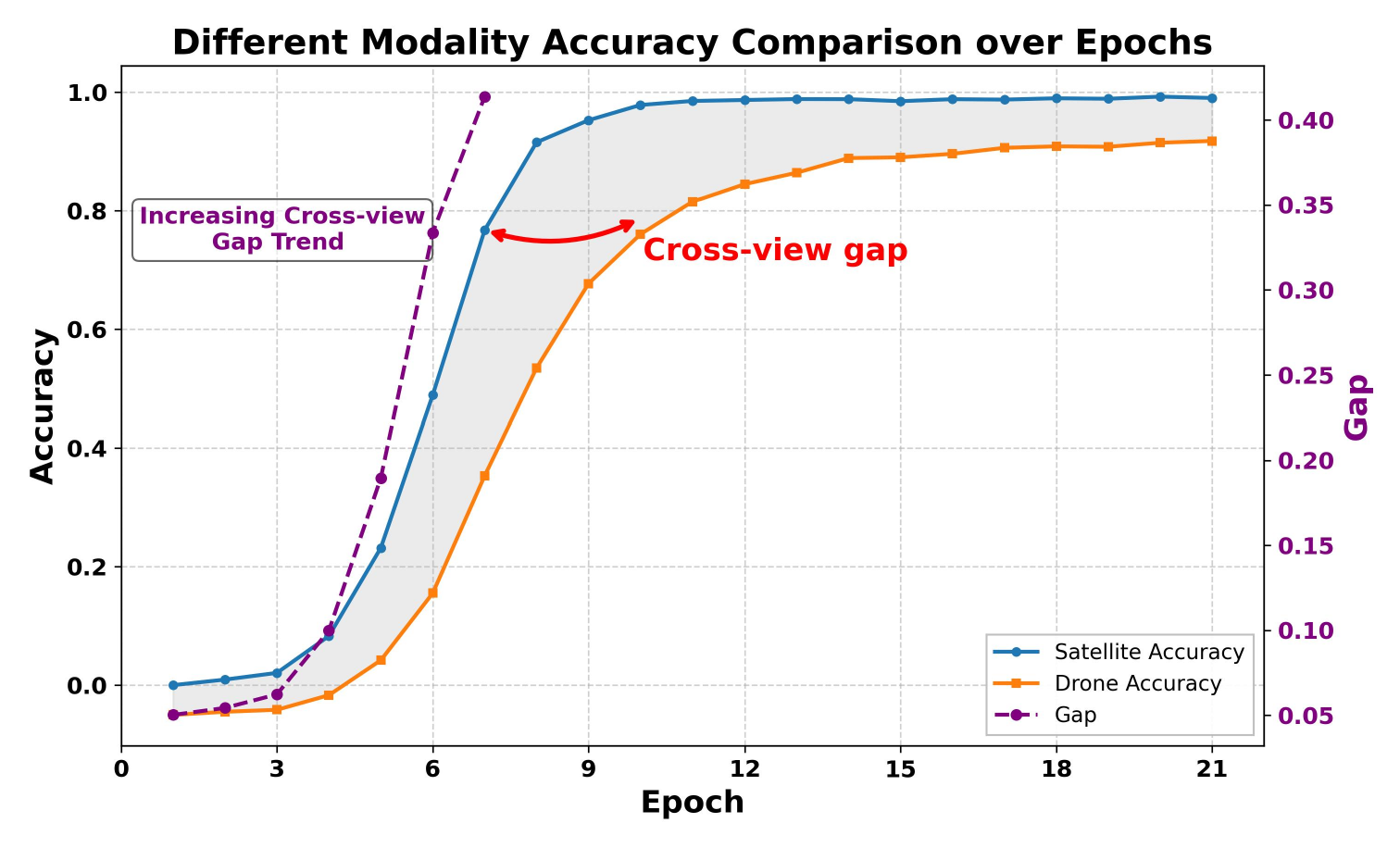}
  \caption{\textbf{Analysis of cross-view performance dynamics. }The figure illustrates the training dynamics of our model, plotting the accuracy for satellite (blue) and drone (orange) domains against training epochs. A noticeable performance discrepancy, or Cross-view Gap (shaded region), emerges where the satellite view consistently outperforms the drone view. Critically, our analysis reveals that this gap is not static but exhibits a clear widening trend during training in the  first several epochs, highlighted by the purple dashed line.}
  \label{accuracy_gap}
  \vspace{-5pt}
\end{figure}

\textbf{Hybrid Loss Function.} The training objectives of HD-CVGL in the first stage (FISD) are threefold, as follows:

\begin{equation}
\label{eq:total_hybrid_loss}
\mathcal{L}_{\text{total}} = \mathcal{L}_{\text{DS}} + \mathcal{L}_{\text{self-dist}} + \mathcal{L}_{\text{metric}}.
\end{equation}

Multi-Level Supervision and Self-Distillation. To ensure that features at all levels of the hierarchy are semantically meaningful, we first apply a standard Cross-Entropy (CE) loss to the logits $z_i$ from each of the $N$ stages. This deep supervision loss is a weighted sum:

\begin{equation}
\label{eq:ds_loss}
\mathcal{L}_{\text{DS}} = \sum_{i=1}^{N} w_i \cdot \mathcal{L}_{\text{CE}}(z_i, y).
\end{equation} 

Further, to regularize the network, we employ intra-model inverse self-distillation. We first define the softened probability distribution for any stage $i$ using a temperature $T$:

\begin{equation}
\label{eq:softmax_temp}
\bm{p}_i(\cdot|T) = \text{Softmax}\left(z_i / T\right).
\end{equation}

Then minimizes the Kullback-Leibler (KL) divergence between teacher distribution ${p}_i$ and the student distribution ${p}_N$:

\begin{equation}
\label{eq:s2d_total}
\mathcal{L}_{\text{self-dist}} = \sum_{i=1}^{N-1} \lambda_i \cdot T^2 \cdot D_{\text{KL}}\left( \bm{p}_i(\cdot|T) \parallel \bm{p}_N(\cdot|T) \right),
\end{equation}
where $D_{\text{KL}}(\cdot \parallel \cdot)$ denotes the KL divergence and $\lambda_i$ are hyper-parameters that weight the contribution of each shallow teacher. This loss regularizes the learning of the final layer.

Refining Embeddings with Symmetric Metric Learning. We introduce a dedicated metric learning objective, $\mathcal{L}_{\text{metric}}$. The goal is to learn a shared, domain-invariant embedding space. Let a batch consist of $B$ pairs of geographically corresponding images $\{(x_k^d, x_k^s)\}_{k=1}^B$, $f_N(x)$ denote the final-stage feature.

First, to enforce intra-domain class separability, we apply the triplet loss with hard sample mining strategy~\cite{peng2024masked,zheng2020university}.

\begin{equation}
\label{eq:triplet}
\mathcal{L}_{\text{triplet}} = \max\left(0, \|f_{N,a} - f_{N,p}\|_2^2 - \|f_{N,a} - f_{N,n}\|_2^2 + m \right).
\end{equation}

Second, to achieve cross-view alignment, we employ a Symmetric InfoNCE Loss~\cite{radford2021learning}. For a drone feature anchor $f_N(x_k^d)$, its corresponding satellite feature $f_N(x_k^s)$ serves as the positive sample. All other satellite features in the batch, $\{f_N(x_l^s)\}_{l \neq k}$, act as negatives. The loss is computed symmetrically, using satellite features as anchors as well. For simplicity, let $f_k^d = f_N(x_k^d)$ and $f_k^s = f_N(x_k^s)$. The Cross-view Symmetric Contrastive (CSC) loss is composed of two symmetric terms: a drone-to-satellite loss ($\mathcal{L}_{d \to s}$) and a satellite-to-drone loss ($\mathcal{L}_{s \to d}$). 

\begin{equation}
\label{eq:csc_main}
\mathcal{L}_{\text{CSC}} = \frac{1}{2} \left( \mathcal{L}_{d \to s} + \mathcal{L}_{s \to d} \right).
\end{equation}
Each directional loss is formulated as an InfoNCE loss. For a batch of $B$ pairs, the drone-to-satellite loss is defined as:
\begin{equation}
\label{eq:csc_d2s}
\mathcal{L}_{d \to s} = -\frac{1}{B}\sum_{k=1}^{B} \log \frac{\exp(\text{sim}(f_k^d, f_k^s) / \tau)}{\sum_{l=1}^{B} \exp(\text{sim}(f_k^d, f_l^s) / \tau)},
\end{equation}
and the satellite-to-drone loss is its symmetric counterpart:
\begin{equation}
\label{eq:csc_s2d}
\mathcal{L}_{s \to d} = -\frac{1}{B}\sum_{k=1}^{B} \log \frac{\exp(\text{sim}(f_k^s, f_k^d) / \tau)}{\sum_{l=1}^{B} \exp(\text{sim}(f_k^s, f_l^d) / \tau)},
\end{equation}
where $\text{sim}(\cdot, \cdot)$ is the cosine similarity and $\tau$ is a temperature parameter. The total deep metric learning objective combines these components, applied to the final feature embeddings:

\begin{equation}
\label{eq:metric_loss_final}
\mathcal{L}_{\text{metric}} = \mathcal{L}_{\text{triplet}}(f_N^d) +  \mathcal{L}_{\text{triplet}}(f_N^s) +  \mathcal{L}_{\text{CSC}}(f_N^d, f_N^s).
\end{equation}

\paragraph{\textbf{Uncertainty-Aware Prediction Alignment}} In cross-view geo-localization, a significant challenge arises from the inherent imbalance and domain discrepancy. Specifically, for each drone-view query, only a single positive satellite sample exists within a large gallery, creating a severe data imbalance. Conventional methods often treat both domains equally, leading to suboptimal feature alignment. As shown in Figure \ref{accuracy_gap}, our observation suggests an asymmetric convergence behavior, where the model may be specializing on the features of the dominant domain. This insight motivates balanced feature learning for robust cross-view matching. 

% % mrm
\begin{figure*}
  \centering
  \includegraphics[width=0.9\linewidth]{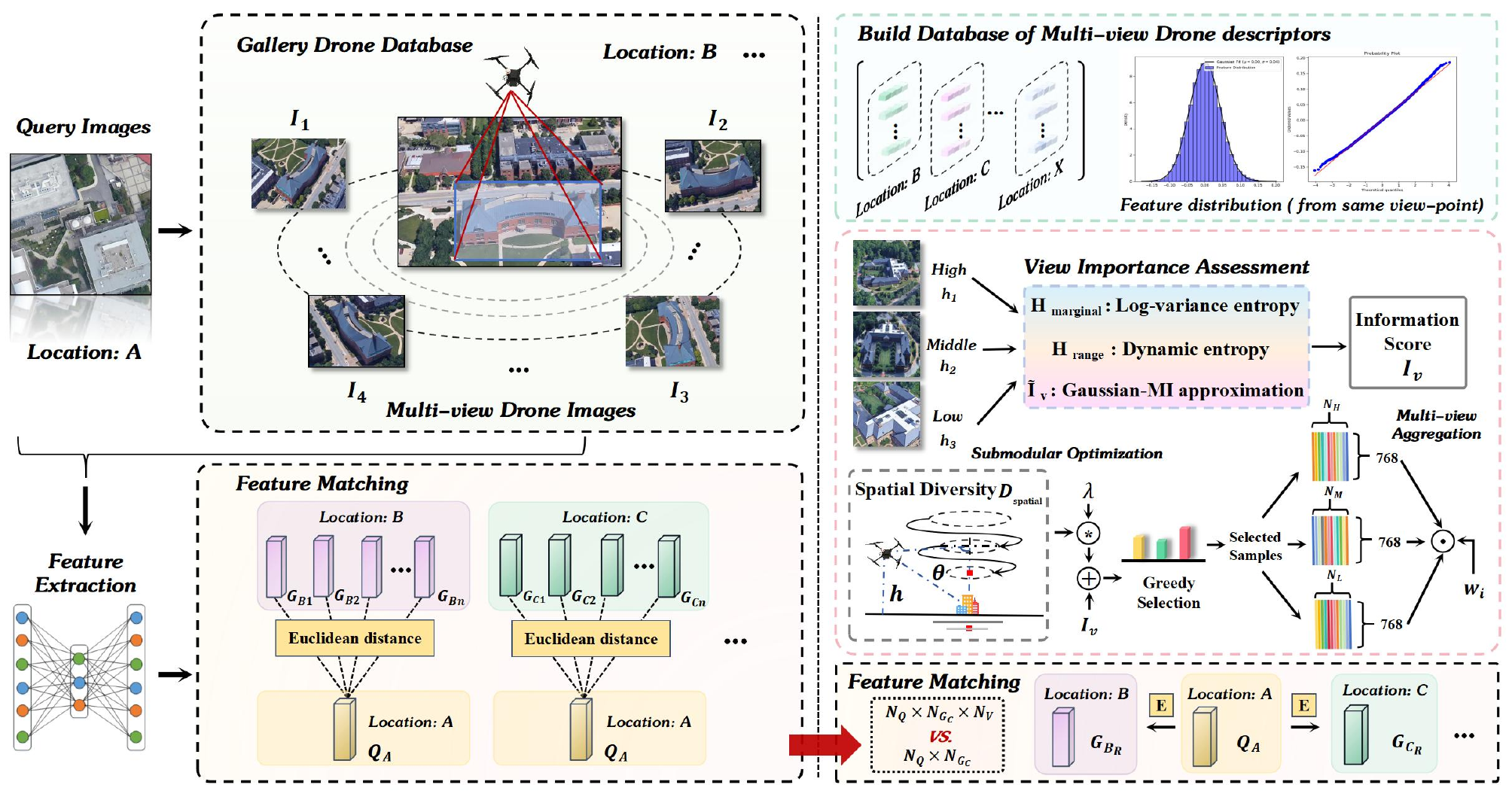}
  \caption{ \textbf{Overview of our Multi-view Selection Refinement Module (MSRM).} On the left, we showcase the feature matching process between multi-view drone images (e.g., from the gallery database) and a satellite image. On the right, the detailed pipeline of MSRM is presented. The process begins with the construction of a multi-view drone descriptor database. As shown, features captured from the same viewpoint are modeled as a Gaussian distribution. The variables $h=[h_1,h_2,h_3]$ and $\theta$ represent the drone's spatial position. Each selected sample is a 768-dimensional vector. The operators $\otimes$, $\odot$, and $\oplus$ denote element-wise multiplication, dot product, and addition, respectively. $E$ represents the Euclidean distance, while $Q_*$$ /$ $G_*$ denote a query $/$ gallery sample.}
  \label{mrm_img}
  \vspace{-5pt}
\end{figure*}

Our approach begins by quantifying the predictive uncertainty of each domain. Inspire by~\cite{ zhou2022uncertainty}, we employ Shannon entropy, a standard measure of uncertainty, calculated from the softmax probabilities derived from the model's output logits. For a given logit vector $\mathbf{z} \in \mathbb{R}^C$ over $C$ classes, the uncertainty $\mathcal{U}$ is defined as:

\begin{equation}
\mathcal{U}(\mathbf{z}) = -\sum_{c=1}^{C} p_c \log p_c, \quad \text{where} \quad p_c = \frac{\exp(z_c)}{\sum_{j=1}^{C} \exp(z_j)},
\end{equation}
here, $p_c$ represents the predicted probability for class $c$, and $z_c$ is the corresponding logit. A higher entropy value signifies greater uncertainty and thus lower confidence in the prediction.

We then dynamically adjust the alignment process based on the relative uncertainty between the drone and satellite branches. We compute the uncertainties $\mathcal{U}_{\text{drone}}$ and $\mathcal{U}_{\text{sat}}$ for the respective logit predictions $\mathbf{z}_{\text{drone}}$ and $\mathbf{z}_{\text{sat}}$. The core of our method is cross-modal self distillation with an adaptive temperature scaling strategy. The temperature $T$ is adjusted based on the uncertainty gap, $\Delta_\mathcal{U}$:

\begin{equation}
\Delta_\mathcal{U} = \mathcal{U}_{\text{drone}} - \mathcal{U}_{\text{sat}},
\end{equation}
\begin{equation}
T = T_0 \times (1 + \sigma(\Delta_\mathcal{U})),
\end{equation}
where $T_0$ is a pre-defined base temperature and $\sigma(\cdot)$ is the sigmoid function. The sigmoid function smoothly maps the unbounded uncertainty gap to a bounded scaling factor in the range (0, 1). This formulation increases the temperature when the drone-view model is more uncertain than the satellite-view model (i.e., $\Delta \mathcal{U} > 0$). This is critical for our self-distillation. When the drone view is ambiguous (e.g., due to view change or occlusions), its model is naturally uncertain. Forcing it to match the satellite's high-confidence prediction would create a conflicting learning signal. Raising the temperature softens the target, mitigating this conflict and providing a more appropriate guidance for the uncertain student.

Finally, we use the adaptive temperature $T$ to guide the alignment between the two domains via a Kullback-Leibler (KL) divergence loss. The satellite branch acts as a "teacher," providing a soft target distribution for the drone "student" branch. The alignment loss, $\mathcal{L}_{\text{align}}$, is formulated as:

\begin{equation}
\mathcal{L}_{\text{align}} = T^2 \cdot \KL\left( \Softmax\left(\frac{\mathbf{z}_{\text{sat}}}{T}\right) \middle\| \Softmax\left(\frac{\mathbf{z}_{\text{drone}}}{T}\right) \right).
\label{eq:align_loss}
\end{equation}

By making the alignment process sensitive to predictive uncertainty, our method fosters a more robust and stable training process, effectively mitigating the challenges posed by data imbalance and domain-specific ambiguity.

\paragraph{\textbf{Cross-distillation training}} The second step of our hierarchical framework is Cross-Distillation Training, a process designed to transfer knowledge from a large foundation model (DINOv2-base~\cite{oquab2023dinov2}) teacher to a lightweight student. Critically, this teacher is not used off-the-shelf; it is first specialized through a parameter-efficient fine-tuning process on the University-1652 dataset, where only the final two Transformer blocks were made trainable. This approach creates an expert teacher that retains general visual knowledge while acquiring high-level semantic understanding specific to CVGL.

To ensure comprehensive knowledge transfer, we distill information at both the feature and logit levels as follows:

\begin{equation}
\mathcal{L}_{\text{logits}} = KL(p^{T},p^{S}).
\end{equation}

\begin{equation}
\mathcal{L}_{\text{feat}} = \underbrace{ \|\phi(F_T) - \phi(F_S) \|_2^2}_{MSE}  +
    \underbrace{ 1 - \frac{\langle \phi(F_T), \phi(F_S) \rangle}{\|\phi(F_T)\|_2 \cdot \|\phi(F_S)\|_2}}_{Cosine \ Similarity},
\end{equation}

\noindent where $\phi$ denotes normalization,  $F_T, F_S$ denote the teacher's and student's final stage output feature, $p^{T}, p^{S}$ are their respective temperature-scaled probability outputs. 

% By distilling both feature semantics and classification logic from the final outputs of a domain-adapted teacher, our student model learns highly relevant representations and decision boundaries for the task.

\subsection{Multi-view Selection Refinement Module (MSRM)}
\label{sec:mrm}

In drone-based visual geo-localization, capturing multiple viewpoints of landmarks is essential for robust matching against satellite references. During data collection, drones systematically capture images at predetermined positions, resulting in a comprehensive multi-view representation. Let $\mathcal{V} = \{v_1, v_2, ..., v_{54}\}$ denote the set of aerial views captured at different heights $h \in \{h_1, h_2, h_3\}$ and azimuth angles $\theta \in \{0°, 20°, ..., 340°\}$. While this dense sampling ensures complete coverage of landmarks, processing all views during inference poses significant computational challenges. 

To address this challenge, as shown in Figure~\ref{mrm_img}, we propose the MSRM, a post-processing technique that intelligently selects an optimal subset $\mathcal{S} \subset \mathcal{V}$ with $|\mathcal{S}| = k \ll |\mathcal{V}|$. The key insight is that not all views contribute equally to geo-localization accuracy: some perspectives capture more distinctive features, while others may be less informative due to occlusions or viewing angles. We formulate this view selection problem within an information-theoretic framework, maximizing the mutual information between landmark identities while ensuring spatial diversity.

\textbf{Theoretical Foundation.} Our approach is grounded in information theory~\cite{vergara2014review}, where the goal is to select views that maximize the mutual information $I(\mathbf{x}_v; y)$ between view features $\mathbf{x}_v$ and landmark labels $y$:

\begin{equation}
I(\mathbf{x}_v; y) = H(\mathbf{x}_v) - H(\mathbf{x}_v|y),
\end{equation}
where $H(\mathbf{x}_v)$ is the differential entropy of view features and $H(\mathbf{x}_v|y)$ is the conditional entropy given the landmark class. Direct computation of mutual information for high-dimensional features is computationally prohibitive~\cite{song2019understanding}. We propose an efficient approximation based on the theoretical connection between Fisher discriminant ratio and mutual information under structured assumptions.

\noindent \textbf{Proposition 1.} Under the assumption that view features follow class-conditional Gaussian distributions with equal covariance, the mutual information can be lower-bounded by:

\begin{equation*}
I(\mathbf{x}_v; y) \geq \frac{1}{2}\log\left(1 + \frac{\sigma^2_{\text{between}}(v)}{\sigma^2_{\text{within}}(v)}\right),
\end{equation*}
where $\sigma^2_{\text{between}}(v)$ and $\sigma^2_{\text{within}}(v)$ are the between-class and within-class variances for view $v$.

This theoretical insight enables us to use computationally efficient statistics as proxies for mutual information while maintaining theoretical rigor.

\textbf{Information-Theoretic View Importance Assessment.} Given extracted multi-view features $\mathbf{X} = \{\mathbf{x}_v \in \mathbb{R}^{D} | v \in \mathcal{V}\}$ for a landmark, the MSRM quantifies each view's information content through three complementary measures grounded in information theory. First, we approximate the marginal entropy of view features through log-variance:

\begin{equation}
H_{\text{marginal}}(v) \approx \frac{1}{2}\log(2\pi e) + \log\left(\frac{1}{D} \sum_{d=1}^{D} \text{std}(\mathbf{x}_v^{(d)})\right).
\end{equation}
This measure captures the information richness of the view, with higher entropy indicating more diverse visual patterns.

Second, we estimate the dynamic entropy through the log-range of feature activations:
\begin{equation}
H_{\text{range}}(v) = \log\left(\frac{1}{D} \sum_{d=1}^{D} [\max(\mathbf{x}_v^{(d)}) - \min(\mathbf{x}_v^{(d)})]\right).
\end{equation}
This complements the variance-based entropy by capturing the span of feature activations, identifying views with strong, distinctive features. 

Most importantly, we compute the Gaussian-MI approximation to measure geo-discriminability:
\begin{equation}
\tilde{I}_v = \frac{1}{2}\log\left(1 + \frac{\sigma^2_{\text{between}}(v)}{\sigma^2_{\text{within}}(v) + \epsilon}\right),
\end{equation}
where the between-class variance quantifies separation between different landmarks:
\begin{equation}
\sigma^2_{\text{between}}(v) = \sum_{c=1}^{C} n_c \|\boldsymbol{\mu}_c^{(v)} - \boldsymbol{\mu}^{(v)}\|^2,
\end{equation}
and the within-class variance measures consistency within each landmark class:
\begin{equation}
\sigma^2_{\text{within}}(v) = \frac{1}{N} \sum_{c=1}^{C} \sum_{i \in \mathcal{I}_c} \|\mathbf{x}_i^{(v)} - \boldsymbol{\mu}_c^{(v)}\|^2,
\end{equation}
here, $n_c$ denotes the number of samples in class $c$, $\boldsymbol{\mu}_c^{(v)}$ is the mean feature for class $c$ in view $v$, $\boldsymbol{\mu}^{(v)}$ is the global mean, and $N$ is the total number of samples. This formulation provides a computationally efficient estimate of mutual information while maintaining theoretical guarantees under Gaussian assumptions. The final information score $I_v$ integrates these measures, where $\hat{(\cdot)}$ denotes min-max normalization:

\begin{equation}
\mathbf{I}_v = \hat{\tilde{I}}_v + \hat{H}_{\text{marginal}}(v) + \hat{H}_{\text{range}}(v).
\end{equation}
 
% and the weights $\alpha = 0.5$, $\beta = 0.3$, $\gamma = 0.2$ reflect the relative importance of class-conditional versus marginal information.

\textbf{Submodular Optimization for Spatial Diversity.} While information scores identify informative views, optimal subset selection must balance information content with spatial coverage. We formulate this as a submodular optimization problem that jointly maximizes information and diversity. 

We model each view's spatial position as $\mathbf{p}_v = (h_v, \theta_v)$, where $h_v$ represents altitude and $\theta_v$ the azimuth angle. The spatial distance between views incorporates both vertical and angular separation:

\begin{equation}
D_{\text{spatial}}(v_i, v_j) = \omega_h \cdot |h_i - h_j| + \omega_{\theta} \cdot d_{\text{circular}}(\theta_i, \theta_j),
\end{equation}
where $d_{\text{circular}}(\theta_i, \theta_j) = \min(|\theta_i - \theta_j|, 360° - |\theta_i - \theta_j|)$ accounts for circular angles, with weights $\omega_h = 2$, $\omega_{\theta} = 1$.

\noindent \textbf{Proposition 2.} The objective function

\begin{equation*}
f(\mathcal{S}) = \sum_{v \in \mathcal{S}} \mathbf{I}_v + \lambda \sum_{v \in \mathcal{S}} \min_{u \in \mathcal{S} \setminus \{v\}} D_{\text{spatial}}(v, u),
\end{equation*}
is submodular, and the greedy algorithm achieves a $(1-1/e)$-approximation to the optimal subset.

The greedy selection iteratively adds views that maximize the marginal gain:
\begin{equation*}
v^* = \arg\max_{v \in \mathcal{V} \setminus \mathcal{S}_t} \left[ \lambda \cdot \mathbf{I}_v + (1-\lambda) \cdot \frac{\min_{s \in \mathcal{S}_t} D_{\text{spatial}}(v, s)}{\max_{u,w \in \mathcal{V}} D_{\text{spatial}}(u, w)} \right],
\end{equation*}
with $\lambda$ balancing information content and spatial diversity.

\textbf{Information-Weighted Multi-view Aggregation.}
After selecting the optimal subset $\mathcal{S}$, we perform information-weighted aggregation that reflects each view's contribution to the mutual information. The aggregation weights are computed using a softmax function over information scores:

\begin{equation}
\mathbf{w}_i = \frac{\exp(\tau \cdot \mathbf{I}_{v_i})}{\sum_{v_j \in \mathcal{S}} \exp(\tau \cdot \mathbf{I}_{v_j})}, \quad \forall v_i \in \mathcal{S},
\end{equation}
where $\tau$ is a temperature parameter controlling the sharpness of the weighting. This exponential weighting amplifies the contribution of high-information views while maintaining differentiability. The refined representation for landmark $l$ is:
\begin{equation}
\mathbf{z}_l = \sum_{v_i \in \mathcal{S}} \mathbf{w}_i \cdot \mathbf{x}_l^{(v_i)} \in \mathbb{R}^D.
\end{equation}

\textbf{Theoretical Analysis and Guarantees.} Our manual information based framework provides the following theoretical guarantees:

\begin{itemize}
  \item Approximation Quality: Under Gaussian assumptions, the approximation error $|I(\mathbf{x}_v; y) - \tilde{I}_v|$ is bounded by $O(\delta^2)$ where $\delta$ measures deviation from Gaussianity.
  % \item Sample Complexity: The number of samples required for reliable MI estimation scales as $O(\sqrt{D}/\epsilon^2)$ for $\epsilon$-accurate estimation, compared to $O(D^2/\epsilon^2)$ for non-parametric estimators.
  \item Computational Efficiency: By reducing the feature matching complexity from $O(N_Q \cdot N_G \cdot |\mathcal{V}|)$ to $O(N_Q \cdot N_G)$, our MSRM achieves a significant speedup. $|\mathcal{V}| \ge 50$ is the number of drone views of one landmark, this corresponds to a $50 \times$ reduction in computational cost, rendering the approach viable for real-time applications.
\end{itemize}

The effectiveness of our approach stems from a principled connection between mutual information and discriminative learning. This connection enables efficient view selection, preserving the most informative perspectives while ensuring comprehensive spatial coverage of drone-view landmarks.

\section{Experiment}
\subsection{Implementation Details}

 We conduct extensive experiments on two prominent drone-based benchmarks that offer complementary characteristics for comprehensive evaluation. University-1652~\cite{zheng2020university} is the first drone-based geo-localization dataset and SUES-200~\cite{zhu2023sues} represents a pioneering benchmark that considers aerial photography captured by drones at different flight heights in the real world. We train our model using a batch size of $64$, where each batch contains $P=8$ different location IDs with $K=4$ samples per ID. This results in $32$ drone images and $32$ satellite images per batch. All images are resized to $224 \times 224$ pixels for both training and testing phases. The model is trained for $60$ epochs using the SGD optimizer, initialized with a learning rate of $0.001$ and incorporating a 5 epoch warm-up phase following~\cite{luo2019bag} to stabilize early-stage gradient dynamics.

% For cross-region evaluation in the unsupervised domain adaptation setting, we implement a contrastive learning fine-tuning strategy, limiting adaptation to a single epoch on the University-1652 dataset to mitigate distribution-specific overfitting. We employ the AutoAugment~\cite{cubuk2019autoaugment} strategy, which applies optimized combinations of augmentation operations including random cropping, translation, horizontal flipping, and color jittering. All experiments are conducted using the PyTorch framework on a single NVIDIA GeForce RTX 3090 GPU with 24GB memory.

% \textbf{Evaluation Protocol.} Following standard practice in cross-view geo-localization~\cite{zheng2020university,zhu2021transgeo}, we evaluate both retrieval accuracy and computational efficiency across all models.

% Retrieval Metrics. We adopt Recall@1 to measure the fraction of queries where the correct result appears in the top position. Average Precision (AP) provides a comprehensive assessment by considering all relevant items in the ranking.

% Efficiency Metrics. We report three efficiency metrics: (1) Parameters: the total number of learnable parameters in millions; (2) FLOPs: floating-point operations computed for a single forward pass; (3) Processing Speed (FPS): in the feature extraction context, the number of drone and satellite images processed per second. Note that we set the batch size to 16 to accommodate the multi-view drone data processing.

% Table 2

\begin{table*}[!t]
\renewcommand{\arraystretch}{0.7}
  \centering
  \caption{Comparison with the recent State-of-the-art Methods on University-1652~\cite{zheng2020university} Dataset. Top-performing  and Second-best Results are Highlighted in \textbf{\textcolor{blue}{Blue}} and \textbf{\textcolor{red}{Red}}. * denotes the efficient model after hierarchical distillation, $^\dagger$ indicates the model after using MSRM post-process.}
  \label{University1652_Results}
  \resizebox{\linewidth}{!}{
    \begin{tabular}{cccccccc}
\toprule[1.5pt]
\toprule

\multirow{2}{*}{\textbf{Method}} &  \multicolumn{3}{c}{}  &  \multicolumn{2}{c}{\textbf{Drone $\rightarrow$ Satellite}}    & \multicolumn{2}{c}{\textbf{Satellite $\rightarrow$ Drone}} \\
&$ \textbf{Parameters} \; \downarrow$  &$ \textbf{FLOPs} \; \downarrow$   &$ \textbf{FPS} \uparrow$      &\textbf{Recall@1} $\uparrow$&\textbf{AP} $\uparrow$  &\textbf{Recall@1} $\uparrow$ &\textbf{AP} $\uparrow$ \\ 
\midrule
LPN~\cite{wang2021each} (Wang et al. 2021)       &62.39 M  &65.39 G    &218    & 77.71  & 80.80   & 90.30 & 78.78     \\
FSRA \cite{dai2021transformer} (Dai et al. 2021)  &53.16 M  &98.05 G    &100    & 85.50  & 87.53   & 89.73 & 84.94     \\
MCCG~\cite{shen2023mccg}  (Shen et al. 2023)     &56.65 M  &51.04 G   &313    & 89.40  & 91.07   & 95.01 & 89.93     \\
MuSe-Net \cite{wang2024multiple} (Wang et al. 2024) &82.90 M  &42.37 G   &405 & 74.48 & 77.83 & 88.02 & 75.10 \\
SCPNet \cite{gao2025semantic}  (Gao et al. 2025)  & -  & -  & -  & 79.96  & 83.04  & 87.33 & 79.87  \\
TriSA  \cite{sun2024tirsa} (Sun et al. 2024)     &51.13 M  &43.18 G    &275    & 90.08  & 91.56   & 96.01 & 90.12    \\
Safe-Net  \cite{lin2024self} (Lin et al. 2024)     & 52.67 M & 24.58 G   & 282    & 86.98  & 88.85  & 91.22 & 86.06 \\
SDPL~\cite{chen2024sdpl}  (Chen et al. 2024)       &42.56 M  &69.71 G    &519    & 90.16  & 91.64   & 93.58 & 89.45     \\
SRLN~\cite{lv2024direction} (Lv et al. 2024) 
& 193.03 M & - & - & 92.70 & 93.77 & 95.14 & 91.97 \\
Sample4Geo~\cite{deuser2023sample4geo} (Deuser et al. 2023) &87.57 M  &90.24 G   &144  & 92.65  & 93.81  & 95.14 & 91.39      \\
CCR~\cite{du2024ccr} (Du et al. 2024) & 156.57 M & 160.61 G &- & 92.54 & 93.78 & 95.15 & 91.80 \\
DAC~\cite{xia2024enhancing} (Xia et al. 2024) & 96.50 M & 90.24 G & 128  & 94.67 & 95.50 & 96.43 & 93.79  \\
\hdashline
MEAN~\cite{chen2025multi}  (Chen et al. 2025)   & \textcolor{blue}{\textbf{36.50 M}} &  \textcolor{blue}{\textbf{26.18 G}}  &  \textcolor{blue}{\textbf{307}} & 93.55 & 94.53 & \textcolor{red}{\textbf{96.01}} & 92.08 \\

\textbf{MobileGeo} $^* \;$ \textbf{(Ours)}  &  \textcolor{red}{\textbf{28.57 M}$_{\;\mathbf{\downarrow}\;\textbf{21.2\%}}$ } & \textcolor{red}{\textbf{4.45 G}$_{\;\mathbf{\downarrow}\;\textbf{5.8}\mathbf{\times}}$ }  & \textcolor{red}{\textbf{1022}$_{\;\mathbf{\uparrow}\;\textbf{3.3}\mathbf{\times}}$ } & \textcolor{blue}{\textbf{93.87}} & \textcolor{blue}{\textbf{94.83}} & \textcolor{blue}{\textbf{95.72}} & \textcolor{blue}{\textbf{92.57}} \\

\textbf{MobileGeo }$^\dagger \;$ \textbf{(Ours w/ Post-process)}  &  \textcolor{red}{\textbf{28.57 M}$_{\;\mathbf{\downarrow}\;\textbf{21.2\%}}$ } & \textcolor{red}{\textbf{4.45 G}$_{\;\mathbf{\downarrow}\;\textbf{5.8}\mathbf{\times}}$ }  & \textcolor{red}{\textbf{1022}$_{\;\mathbf{\uparrow}\;\textbf{3.3}\mathbf{\times}}$ } & \textcolor{red}{\textbf{97.15$_{\;\mathbf{\uparrow}\;\textbf{3.30 \%}}$}} & \textcolor{red}{\textbf{97.50$_{\;\mathbf{\uparrow}\;\textbf{2.97 \%}}$}} & 95.58 & \textcolor{red}{\textbf{96.27$_{\;\mathbf{\uparrow}\;\textbf{4.19 \%}}$}} \\
      \bottomrule
      \bottomrule[1.5pt]
    \end{tabular}
    }
    
\end{table*}

% Table 3
% Table 4
\renewcommand{\arraystretch}{1} % Adjust the row height
\begin{table*}[!t]
\centering
\small
\caption{Comparisons between the proposed method and state-of-the-art methods in unsupervised domain adaption evaluation (from University-1652 to SUES-200) on Satellite$\rightarrow$Drone. }
\label{uda-sd}
\resizebox{\textwidth}{!}{
\begin{tabular}{cccccccccccc}
\toprule[1.5pt]
\toprule
 &  &  &  & \multicolumn{8}{c}{\textbf{Satellite$\rightarrow$Drone}} \\ \cline{5-12} 
 
 &  &  &  & \multicolumn{2}{c}{\textbf{150m}} & \multicolumn{2}{c}{\textbf{200m}} & \multicolumn{2}{c}{\textbf{250m}} & \multicolumn{2}{c}{\textbf{300m}} \\ 
 
\multirow{-3}{*}{\textbf{Model}} &\multirow{-1}{*}{$ \textbf{Parameters} \; \downarrow$} & \multirow{-1}{*}{$ \textbf{FLOPs} \; \downarrow$} & \multirow{-1}{*}{$ \textbf{FPS} \; \uparrow$} &\textbf{R@1} $\uparrow$ &\textbf{AP} $\uparrow$ & \textbf{R@1} $\uparrow$ &\textbf{AP} $\uparrow$ & \textbf{R@1} $\uparrow$  & \textbf{AP} $\uparrow$ & \textbf{R@1} $\uparrow$ & \textbf{AP} $\uparrow$ \\ 

\midrule

MCCG~\cite{shen2023mccg}  (Shen et al. 2023) 
&56.65 M   &51.04 G & 313 & 61.25 & 53.51 & 82.50 & 67.06 & 81.25 & 74.99 & 87.50 & 80.20   \\

Sample4Geo~\cite{deuser2023sample4geo} (Deuser et al. 2023)
& 87.57 M &90.24 G & 144 & 83.75 & 73.83 & 91.25 & 83.42 &93.75 & 89.07 & 93.75 & 90.66   \\

DAC\cite{xia2024enhancing} (Xia et al. 2024)       
& 96.50 M & 90.24 G & 128 & 87.50 & 79.87 &96.25 &88.98  &95.00 & 92.81 &96.25 & 94.00  \\

MEAN~\cite{chen2025multi}  (Chen et al. 2025)
 & \textcolor{blue}{\textbf{36.50 M}} &  \textcolor{blue}{\textbf{26.18 G}}  &  \textcolor{blue}{\textbf{307}}  &  \textcolor{blue}{\textbf{91.25}}   &  \textcolor{blue}{\textbf{81.50}}   &  \textcolor{blue}{\textbf{96.25}}   &  \textcolor{blue}{\textbf{89.55}}  &  \textcolor{blue}{\textbf{95.00}} &  \textcolor{blue}{\textbf{92.36}} &  \textcolor{blue}{\textbf{96.25}}  &  \textcolor{blue}{\textbf{94.32}}  \\
\midrule

\textbf{MobileGeo} $^* \;$ \textbf{(Ours)}  &  \textcolor{red}{\textbf{28.57 M}$_{\;\mathbf{\downarrow}\;\textbf{21.2\%}}$ } & \textcolor{red}{\textbf{4.45 G}$_{\;\mathbf{\downarrow}\;\textbf{5.8}\mathbf{\times}}$ }  & \textcolor{red}{\textbf{1022}$_{\;\mathbf{\uparrow}\;\textbf{3.3}\mathbf{\times}}$ }
& \textcolor{red}{\textbf{92.50}} & \textcolor{red}{\textbf{83.81}} & \textcolor{red}{\textbf{97.50}} & \textcolor{red}{\textbf{91.75}} & \textcolor{red}{\textbf{98.75}} & \textcolor{red}{\textbf{94.59}} & \textcolor{red}{\textbf{97.50}} & \textcolor{red}{\textbf{96.04}}\\

\bottomrule
\bottomrule[1.5pt]
\end{tabular}}
\end{table*}

\subsection{Comparison with State-of-the-Art Methods}

\textbf{Superior Efficiency.} As presented in Table~\ref{University1652_Results}, our core model, MobileGeo, exhibits high computational efficiency. With only 28.57M parameters and an exceptionally low 4.45G FLOPs, it is by far the most lightweight and computationally inexpensive model among all compared methods. To put this in perspective, compared to the recent efficient model MEAN~\cite{chen2025multi}, MobileGeo reduces FLOPs by a remarkable factor of 5.8$\times$ and parameters by 21.2\%. This optimization directly translates to a massive 3.3$\times$ increase in FPS, reaching 1022 images/second, a critical capability for real-time deployment. 

% Furthermore, MobileGeo achieves a highly competitive Recall@1 of 97.15\% on the Drone$\rightarrow$Satellite task, outperforming many heavier models such as Sample4Geo~\cite{deuser2023sample4geo} and SRLN~\cite{lv2024direction}.

\textbf{State-of-the-Art Accuracy.} Building upon this highly efficient foundation, our full model, MobileGeo$^\dagger$, incorporates the MSRM as a post-processing step achieves an impressive 97.15\% Recall@1 and 97.50\% AP in the primary Drone$\rightarrow$Satellite retrieval task. This represents a substantial absolute improvement of 3.30\% in R@1 over our efficient baseline and significantly surpasses the previous best-performing method, DAC~\cite{xia2024enhancing}, all while operating with over 20$\times$ fewer FLOPs (4.45G vs. 90.24G).

% that does not add any computational overhead during feature extraction. This enhancement pushes the performance to a new state-of-the-art. In the primary Drone$\rightarrow$Satellite retrieval task, MobileGeo$^\dagger$ 

% In summary, the results in Table~\ref{University1652_Results} unequivocally show that MobileGeo successfully achieves a trade-off between accuracy and efficiency in cross-view geo-localization, making it a highly practical and effective solution for real-world applications.

\subsection{Unsupervised Domain Adaptation Results}

To rigorously assess the generalization capabilities of our model, we conducted zero-shot evaluations by training on the \textit{University-1652} and directly testing on the \textit{SUES-200} without any fine-tuning. This challenging setting simulates real-world deployment where models must handle unseen data domains. 

% As demonstrated in Table~\ref{uda-ds}, MobileGeo exhibits exceptional generalization in the Drone$\rightarrow$Satellite task. While MEAN~\cite{chen2025multi} shows strong performance at 150m, our model consistently outperforms all state-of-the-art methods at 200m, 250m, and 300m. Notably, at the 300m range, MobileGeo achieves a top-ranking R@1 score of 97.87, surpassing the next-best competitor by a significant 3.24 percentage points. It is crucial to highlight that MobileGeo utilizes only 4.45G FLOPs, compared to the 26.18G FLOPs required by MEAN. 

As demonstrated in Table~\ref{uda-sd}, MobileGeo exhibits exceptional generalization in the Satellite$\rightarrow$Drone task. In this scenario, our model unequivocally achieves the best performance across all evaluation altitudes. For instance, it surpasses the strong baseline DAC~\cite{xia2024enhancing} by 3.75 percentage points in R@1 at the 250m altitude.

\subsection{Multi-weather drone imagery degradation Results}

We conducted extensive experiments under various environmental degradations, and as shown in Table~\ref{weather}, the model maintains high accuracy despite severely compromised visual quality in drone imagery. In the Drone$\rightarrow$Satellite retrieval task, our proposed MobileGeo establishes a new state-of-the-art across all tested conditions. From normal weather to the most severe degradations like darkness and combined fog with rain, MobileGeo consistently achieves the highest Recall@1 and AP scores. For instance, under dark conditions, it outperforms the next-best method, MEAN~\cite{chen2025multi}, by a substantial margin of 5.37 percentage points in R@1 (93.27\% vs. 87.90\%). In the more challenging Satellite$\rightarrow$Drone task, MobileGeo continues to show highly competitive performance.

\subsection{Anti-offset Generalization Results}
\label{sec:anti_offset}

% results1
% \begin{figure}
%   \centering
%   \includegraphics[width=\linewidth]{images/results1.pdf}
%   \caption{\textbf{Top-5 retrieval results on the University-1652 dataset.} Green boxes denote correct matches and red boxes denote incorrect matches.}
%   \label{results1}
%   \vspace{-5pt}
% \end{figure}

% results2
% \begin{figure}
%   \centering
%   \includegraphics[width=\linewidth]{images/results2.pdf}
%   \caption{\textbf{Top-5 retrieval results on the SUES-200 dataset.} Green boxes denote correct matches and red boxes denote incorrect matches.}
%   \label{results2}
%   \vspace{-5pt}
% \end{figure}

In real-world CVGL, the captured drone image is often not perfectly centered with its corresponding satellite-view image. This spatial misalignment can be caused by variations in camera angle or differences in viewpoint. A robust model must be able to generalize well despite such spatial offsets.

We adopt the evaluation protocol popularized by SDPL~\cite{chen2024sdpl}. The mapping from a desired shift $(\Delta x, \Delta y)$ to the required (left, top, right, bottom) padding tuple is as follows: top-left $(-\text{P},-\text{P})$, top-right $(-\text{P},+\text{P})$, bottom-left $(+\text{P},-\text{P})$, and bottom-right $(+\text{P},+\text{P})$. The comprehensive results are presented in Table~\ref{offset}. Under the most severe shift of $(-60,-60)$ pixels, where other models experience a significant performance collapse, MobileGeo maintains an exceptional Rank-1 accuracy of 93.86\%. This represents a massive improvement of +16.92\% over the second-best method, SDPL.

% This highlights its ability to learn features that are inherently invariant to spatial translations. 

% This anti-offset generalization capability, coupled with its significantly lower computational cost (4.45 GFLOPs) compared to competitors like FSRA (98.05 GFLOPs) and SDPL (69.71 GFLOPs), underscores the practical value and efficiency of our method for real-world deployment.

% Table 5
\renewcommand{\arraystretch}{1}
\begin{table}[!t]
\renewcommand{\arraystretch}{0.5}
\caption{Comparison with state-of-the-art results under multi-weather conditions on the university-1652 dataset.}
\label{weather}
\resizebox{\linewidth}{!}{
\begin{tabular}{cccccc}
\toprule[1.5pt]
\toprule
\multirow{2}{*}{\textbf{Method}}  &  &  \multicolumn{2}{c}{\textbf{Drone$\rightarrow$Satellite}}   & \multicolumn{2}{c}{\textbf{Satellite$\rightarrow$Drone}} \\

& \textbf{FLOPs}$_{\;\mathbf{\downarrow}}$ & \textbf{Recall@1}$_{\;\mathbf{\uparrow}}$ & \textbf{AP}$_{\;\mathbf{\uparrow}}$ & \textbf{Recall@1}$_{\;\mathbf{\uparrow}}$ & \textbf{AP}$_{\;\mathbf{\uparrow}}$   \\ 

\midrule
% \multicolumn{6}{c}{\textbf{(a) Normal}}  \\
% \midrule 
% LPN~\cite{wang2021each}               & 65.39\; G  &74.33  &77.60  &87.02 &75.19 \\
% MuSeNet~\cite{wang2024multiple}       & 42.37\; G  &74.48  &77.83  &88.02 &75.10  \\
% Sample4Geo\cite{deuser2023sample4geo} & 90.24\; G  &90.55  &92.18  &95.86 &89.86  \\
% MEAN\cite{chen2025multi}              & \textcolor{blue}{\textbf{26.18\; G}}  &\textcolor{blue}{\textbf{90.81}}  &\textcolor{blue}{\textbf{92.32}}  &\textcolor{red}{\textbf{96.58}} &\textcolor{blue}{\textbf{89.93}}  \\
% \midrule
% \textbf{MobileGeo} $^* \;$ \textbf{(Ours)}     & \textcolor{red}{\textbf{4.45 G}$_{\; \mathbf{\downarrow} \; \textbf{5.8}  \mathbf{\times}}$ }  & \textcolor{red}{\textbf{93.87}}  & \textcolor{red}{\textbf{94.83}} & \textcolor{blue}{\textbf{95.72}} & \textcolor{red}{\textbf{92.57}}  \\

\midrule
\multicolumn{6}{c}{\textbf{\faSmog~~(a) Fog}}  \\
\midrule 
LPN~\cite{wang2021each}               & 65.39\; G  & 69.31 & 72.95 & 86.16 & 71.34 \\
MuSeNet~\cite{wang2024multiple}       & 42.37\; G  & 69.47 & 73.24 & 87.87 & 69.85 \\
Sample4Geo\cite{deuser2023sample4geo} & 90.24\; G  & 89.72 & 91.48 & 95.72 & 88.95 \\
MEAN\cite{chen2025multi}              & \textcolor{blue}{\textbf{26.18\; G}}  &\textcolor{blue}{\textbf{90.97}}  &\textcolor{blue}{\textbf{92.52}}  &\textcolor{red}{\textbf{96.00}} &\textcolor{blue}{\textbf{89.49}}  \\
\midrule
\textbf{MobileGeo} $^* \;$ \textbf{(Ours)} & \textcolor{red}{\textbf{4.45 G}$_{\; \mathbf{\downarrow} \; \textbf{5.8}  \mathbf{\times}}$ }  & \textcolor{red}{\textbf{92.95}}  & \textcolor{red}{\textbf{94.08}} & \textcolor{blue}{\textbf{95.72}} & \textcolor{red}{\textbf{91.17}}  \\

\midrule
\multicolumn{6}{c}{\textbf{\faCloudRain~~(b) Rain}}  \\
\midrule
LPN~\cite{wang2021each}               & 65.39\; G  & 67.96 & 71.72 & 83.88 & 69.49\\
MuSeNet~\cite{wang2024multiple}       & 42.37\; G  & 70.55 & 74.14 & 87.73 & 71.12 \\
Sample4Geo~\cite{deuser2023sample4geo} & 90.24\; G  & 85.89 & 88.11 & 94.44 & 85.71 \\
MEAN~\cite{chen2025multi}              & \textcolor{blue}{\textbf{26.18\; G}}  &\textcolor{blue}{\textbf{88.19}}  &\textcolor{blue}{\textbf{90.05}}  &\textcolor{red}{\textbf{95.15}} &\textcolor{blue}{\textbf{88.87}}  \\
\midrule
\textbf{MobileGeo} $^* \;$ \textbf{(Ours)} & \textcolor{red}{\textbf{4.45 G}$_{\; \mathbf{\downarrow} \; \textbf{5.8}  \mathbf{\times}}$ }  & \textcolor{red}{\textbf{91.26}}  & \textcolor{red}{\textbf{92.61}} & \textcolor{blue}{\textbf{94.58}} & \textcolor{red}{\textbf{87.22}}  \\

% \midrule
% \multicolumn{6}{c}{\textbf{(d) Fog+Rain}}  \\
% \midrule
% LPN~\cite{wang2021each}               & 65.39\; G  & 64.51 & 68.52  & 84.59 & 66.28 \\
% MuSeNet~\cite{wang2024multiple}       & 42.37\; G  & 65.59 & 69.64  & 85.02 & 67.78 \\
% Sample4Geo~\cite{deuser2023sample4geo} & 90.24\; G  & 85.88 & 88.16  & 93.44 & 85.27\\
% MEAN~\cite{chen2025multi}              & \textcolor{blue}{\textbf{26.18\; G}}  &\textcolor{blue}{\textbf{86.75}}  &\textcolor{blue}{\textbf{88.84}}  &\textcolor{blue}{\textbf{93.58}} &\textcolor{red}{\textbf{86.91}}  \\
% \midrule
% \textbf{MobileGeo} $^* \;$ \textbf{(Ours)}  & \textcolor{red}{\textbf{4.45 G}$_{\; \mathbf{\downarrow} \; \textbf{5.8}  \mathbf{\times}}$ }  & \textcolor{red}{\textbf{90.68}}  & \textcolor{red}{\textbf{92.11}} & \textcolor{red}{\textbf{94.86}} & \textcolor{blue}{\textbf{85.67}}  \\

\midrule
\multicolumn{6}{c}{\textbf{\faMoon~~(c) Dark}}  \\
\midrule
LPN~\cite{wang2021each}               & 65.39\; G  & 53.68 & 58.10 & 82.88 & 52.05 \\
MuSeNet~\cite{wang2024multiple}       & 42.37\; G  & 53.85 & 58.49 & 80.74 & 53.01 \\
Sample4Geo\cite{deuser2023sample4geo} & 90.24\; G  & 87.90 & 89.87 & 96.01 & 87.06 \\
MEAN\cite{chen2025multi}              & \textcolor{blue}{\textbf{26.18\; G}}  &\textcolor{blue}{\textbf{87.90}}  &\textcolor{blue}{\textbf{89.87}}  &\textcolor{red}{\textbf{96.29}} &\textcolor{blue}{\textbf{89.87}}  \\
\midrule
\textbf{MobileGeo} $^* \;$ \textbf{(Ours)}   & \textcolor{red}{\textbf{4.45 G}$_{\; \mathbf{\downarrow} \; \textbf{5.8}  \mathbf{\times}}$ }  & \textcolor{red}{\textbf{93.27}}  & \textcolor{red}{\textbf{94.34}} & \textcolor{blue}{\textbf{95.44}} & \textcolor{red}{\textbf{89.95}}  \\

\midrule
\multicolumn{6}{c}{\textbf{\faWind~~(d) Wind}}  \\
\midrule
LPN~\cite{wang2021each}               & 65.39\; G  & 66.46 & 70.35 & 84.14 & 67.35 \\
MuSeNet~\cite{wang2024multiple}       & 42.37\; G  & 69.45 & 73.22 & 86.31 & 70.03 \\
Sample4Geo\cite{deuser2023sample4geo} & 90.24\; G  & 83.39 & 89.51 & 95.29 & 87.06 \\
MEAN\cite{chen2025multi}              & \textcolor{blue}{\textbf{26.18\; G}}  &\textcolor{blue}{\textbf{89.27}}  &\textcolor{blue}{\textbf{91.01}}  &\textcolor{blue}{\textbf{95.44}} &\textcolor{blue}{\textbf{86.05}}  \\
\midrule
\textbf{MobileGeo} $^* \;$ \textbf{(Ours)}   & \textcolor{red}{\textbf{4.45 G}$_{\; \mathbf{\downarrow} \; \textbf{5.8}  \mathbf{\times}}$ }  & \textcolor{red}{\textbf{93.45}}  & \textcolor{red}{\textbf{94.48}} & \textcolor{red}{\textbf{95.58}} & \textcolor{red}{\textbf{91.61}}  \\

% \midrule
% \multicolumn{6}{c}{\textbf{(g) Wind}}  \\
% \midrule
% LPN~\cite{wang2021each}               & 65.39\; G  & 66.46 & 70.35 & 84.14 & 67.35\\
% MuSeNet~\cite{wang2024multiple}       & 42.37\; G  & 69.45 & 73.22 & 86.31 & 70.03 \\
% Sample4Geo\cite{deuser2023sample4geo} & 90.24\; G  & 83.39 & 89.51 & 95.29 & 87.06 \\
% MEAN\cite{chen2025multi}              & \textcolor{blue}{\textbf{26.18\; G}}  &\textcolor{blue}{\textbf{89.27}}  &\textcolor{blue}{\textbf{91.01}}  &\textcolor{blue}{\textbf{95.44}} &\textcolor{blue}{\textbf{86.05}}  \\
% \midrule
% \textbf{MobileGeo} $^* \;$ \textbf{(Ours)}  & \textcolor{red}{\textbf{4.45 G}${_\; \mathbf{\downarrow} \; \textbf{5.8}  \mathbf{\times}}$ }  & \textcolor{red}{\textbf{93.45}}  & \textcolor{red}{\textbf{94.48}} & \textcolor{red}{\textbf{95.58}} & \textcolor{red}{\textbf{91.61}}  \\

\bottomrule
\bottomrule[1.5pt]
\end{tabular}}
\vspace{-4pt}
\end{table}

\begin{figure}
  \centering
  \includegraphics[width=0.8\linewidth]{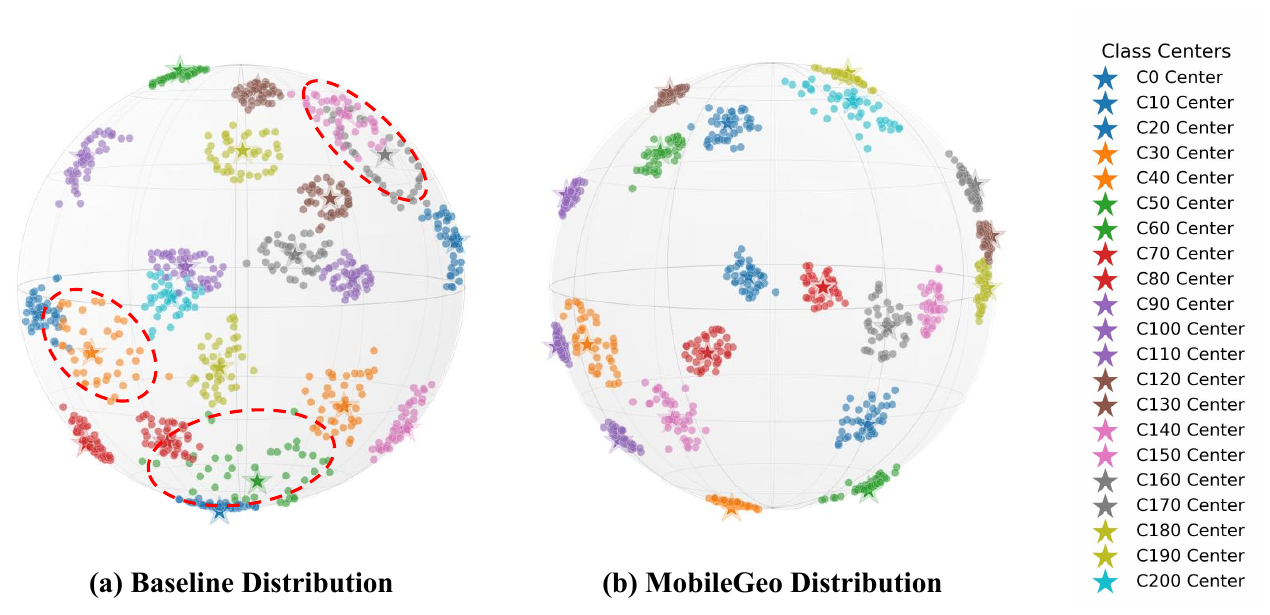}
  \caption{\textbf{t-SNE~\cite{maaten2008visualizing} visualization of drone-view feature embeddings in 3D feature space}, projected onto a spherical surface for better observation. We selected 20 locations with 40 samples per location.}
  \label{t-sne}
  \vspace{-5pt}
\end{figure}

\begin{figure}
  \centering
  \includegraphics[width=0.8\linewidth]{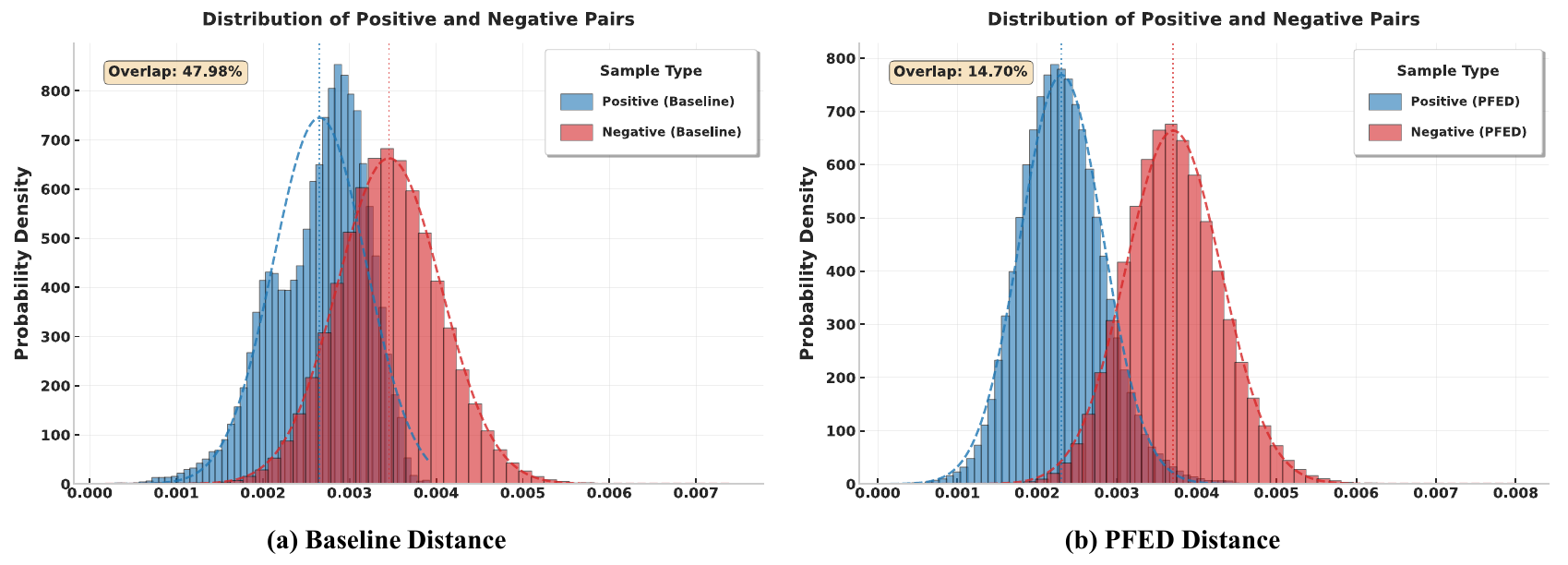}
  \caption{\textbf{Distance distribution of all positive and negative sample pairs in the test set.} Blue and red represent the distance distributions of positive (intra-class) and negative (inter-class) sample pairs, respectively.}
  \label{distance}
  \vspace{-5pt}
\end{figure}

\subsection{Ablation Studies}
\label{sec:ablation}

\begin{table*}[ht]
\renewcommand{\arraystretch}{0.2}  % 保持你的行高压缩
\centering
\small  % 建议保留，数字更清晰（可换 \footnotesize 如果还想更小）

\caption{Comparison with state-of-the-art results on the University-1652 dataset with different shifting sizes of query images DURING INFERENCE. We report the retrieval results and performance improvement of our MobileGeo in five padding patterns.}
\label{offset}

\setlength{\tabcolsep}{0pt}
\begin{tabular*}{0.95\textwidth}{@{\extracolsep{\fill}} 
    *{9}{c}   % ← 9 centered columns
}
\toprule[1.5pt]
\multirow{2}{*}{\textbf{Padding Pixel}}  
& \multicolumn{2}{c}{\textbf{FSRA (98.05 G)}} 
& \multicolumn{2}{c}{\textcolor{blue}{\textbf{LPN (65.39 G)}}}  
& \multicolumn{2}{c}{\textbf{SDPL (69.71 G)}} 
& \multicolumn{2}{c}{\textcolor{red}{\textbf{MobileGeo (4.45 G)}}}\\ 
\cmidrule(lr){2-3} \cmidrule(lr){4-5} \cmidrule(lr){6-7} \cmidrule(lr){8-9} 
& Recall@1 & AP & Recall@1 & AP & Recall@1 & AP & Recall@1 & AP \\
% \midrule
% \midrule

% (0,0)   & \textcolor{blue}{\textbf{86.41}} & \textcolor{blue}{\textbf{88.34}} & 77.99 & 80.84 & 85.19 & 87.43 & \textcolor{red}{\textbf{93.87 (+7.46)}} & \textcolor{red}{\textbf{94.83 (+6.49)}} \\
% (+20,0) & \textcolor{blue}{\textbf{85.51}} & \textcolor{blue}{\textbf{87.59}} & 76.64 & 79.62 & 84.42 & 86.78 & \textcolor{red}{\textbf{93.19 (+7.68)}} & \textcolor{red}{\textbf{94.28 (+6.69)}} \\
% (+40,0) & \textcolor{blue}{\textbf{82.77}} & \textcolor{blue}{\textbf{85.30}} & 72.94 & 76.36 & 82.46 & 85.13 & \textcolor{red}{\textbf{90.84 (+8.07)}} & \textcolor{red}{\textbf{92.34 (+7.04)}} \\
% (+60,0) & 77.95 & 81.18 & 66.85 & 70.91 & \textcolor{blue}{\textbf{79.22}} & \textcolor{blue}{\textbf{82.43}} & \textcolor{red}{\textbf{85.17 (+5.95)}} & \textcolor{red}{\textbf{87.59 (+5.16)}} \\

\midrule
\midrule

(-20,-20) & 84.35 & 86.62 & 76.40 & 79.42 &  \textcolor{blue}{\textbf{84.39}} &  \textcolor{blue}{\textbf{86.76}} & \textcolor{red}{\textbf{93.87 (+9.48)}} & \textcolor{red}{\textbf{94.83(+8.07)}}\\
(-40,-40) & 78.10 & 81.24 & 70.27 & 74.03 &  \textcolor{blue}{\textbf{81.75}} &  \textcolor{blue}{\textbf{84.55}} & \textcolor{red}{\textbf{93.88 (+12.13)}} & \textcolor{red}{\textbf{94.84 (+10.29)}}\\
(-60,-60) & 67.73 & 71.97 & 59.56 & 64.34 &  \textcolor{blue}{\textbf{76.94}} &  \textcolor{blue}{\textbf{80.46}} & \textcolor{red}{\textbf{93.86 (+16.92)}} & \textcolor{red}{\textbf{94.82 (+14.36)}}\\

\midrule
\midrule

(+20,-20) & 84.23 & 86.55 & 76.34 & 79.37 & \textcolor{blue}{\textbf{84.32}} & \textcolor{blue}{\textbf{86.72}} & \textcolor{red}{\textbf{93.19 (+8.87)}} & \textcolor{red}{\textbf{94.28 (+7.56)}}\\
(+40,-40) & 77.90 & 81.09 & 70.36 & 74.10 & \textcolor{blue}{\textbf{81.62}} & \textcolor{blue}{\textbf{84.46}} & \textcolor{red}{\textbf{90.85 (+9.62)}} & \textcolor{red}{\textbf{92.35 (+8.31)}} \\
(+60,-60) & 67.29 & 71.62 & 59.61 & 64.42 & \textcolor{blue}{\textbf{76.80}} & \textcolor{blue}{\textbf{80.38}} & \textcolor{red}{\textbf{85.16 (+8.36)}} & \textcolor{red}{\textbf{87.58 (+7.20)}} \\

\midrule
\midrule
(-20,+20) & \textcolor{blue}{\textbf{83.46}} & \textcolor{blue}{\textbf{85.85}} & 74.74 & 77.92 & 82.95 & 85.49 & \textcolor{red}{\textbf{92.72 (+9.26)}} & \textcolor{red}{\textbf{93.88 (+8.03)}}\\
(-40,+40) & 74.47 & 78.05 & 65.13 & 69.32 & \textcolor{blue}{\textbf{77.00}} & \textcolor{blue}{\textbf{80.46}} & \textcolor{red}{\textbf{86.62 (+9.62)}} & \textcolor{red}{\textbf{88.77 (+8.31)}}\\
(-60,+60) & 58.05 & 63.27 & 50.19 & 55.56 & \textcolor{blue}{\textbf{66.87}} & \textcolor{blue}{\textbf{71.71}} & \textcolor{red}{\textbf{72.36 (+5.49)}} & \textcolor{red}{\textbf{76.25 (+4.54)}}\\

\bottomrule
\bottomrule[1.5pt]

\end{tabular*}
\vspace{-8pt}
\end{table*}

\begin{figure}
  \centering
  \includegraphics[width=0.9\linewidth]{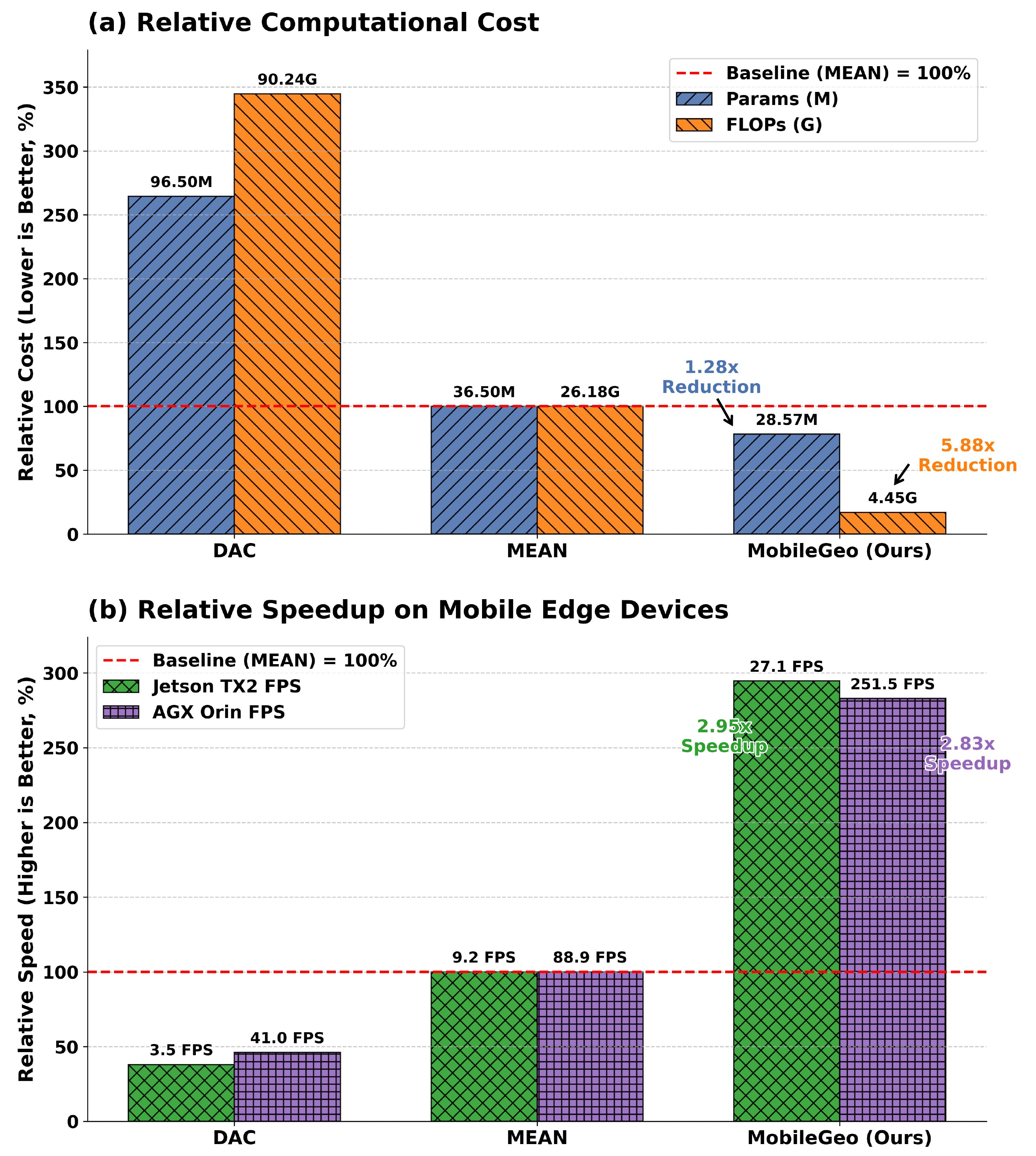}
  \caption{The plots show (a) reduced computational cost (Params, FLOPs) and (b) increased inference speed (FPS on TX2, Orin) for MobileGeo compared to DAC and MEAN, with MEAN set as 100\%. MobileGeo (Ours) outperforms baselines in computational efficiency and edge device speed by a large margin.}
  \label{tx2}
  \vspace{-5pt}
\end{figure}

% onboard
\begin{figure}
  \centering
  \includegraphics[width=0.9\linewidth]{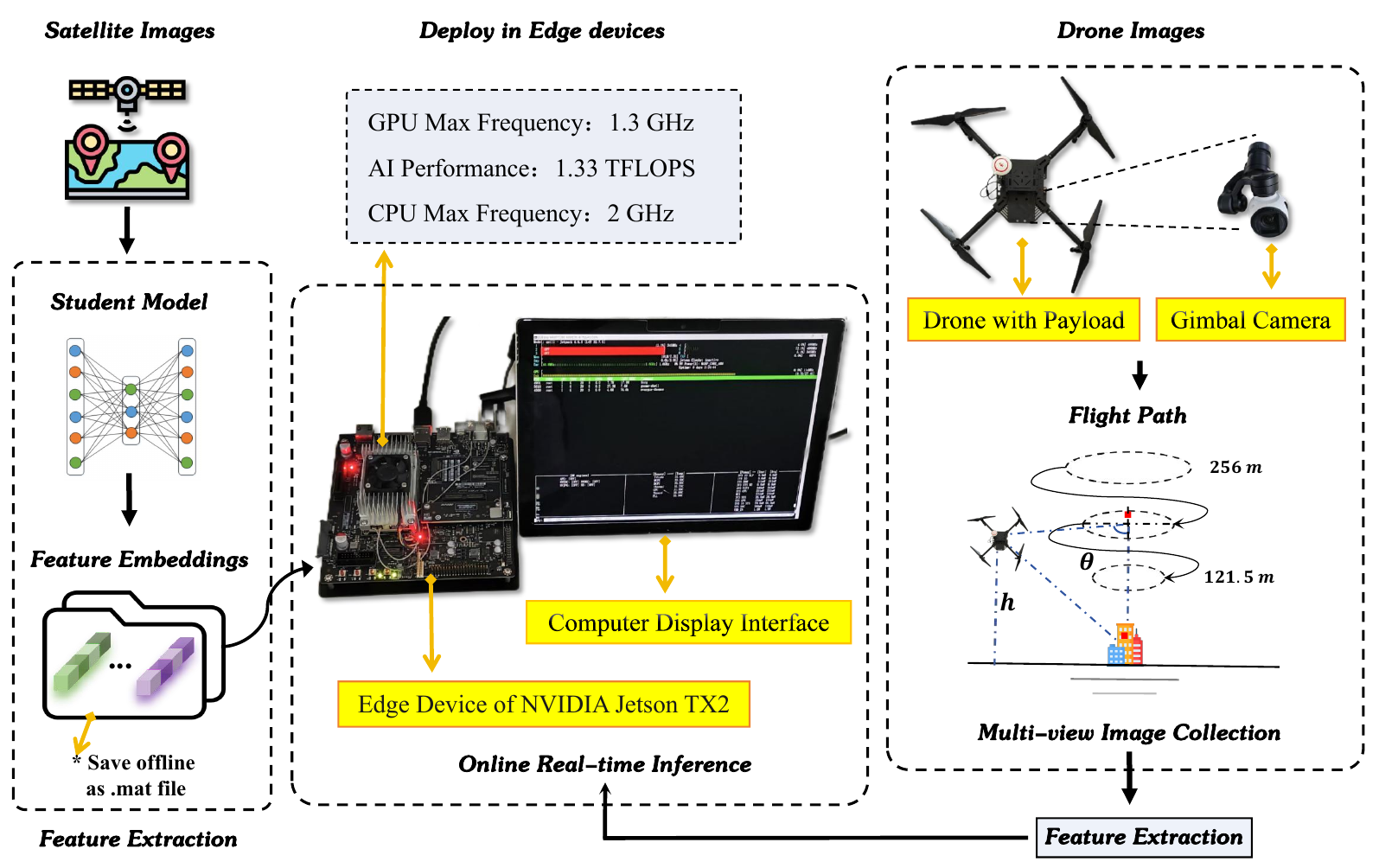}
  \caption{\textbf{Deployment pipeline of cross-view geo-localization model on edge devices.} Satellite images are offline encoded into feature embeddings via our model and stored as $.mat$ files on edge devices. drone captures multi-view images in real-time, which are processed through feature extraction and matched with satellite feature embeddings to obtain current GPS information.}
  \label{onboard}
  \vspace{-5pt}
\end{figure}

\textbf{Effectiveness of the Training-Phase Framework.} The core objective of our training strategy is to distill rich, view-invariant knowledge into a single, efficient network, avoiding the need for multi-branch architectures during inference. As summarized in Table~\ref{ab}, our analysis starts with a baseline model (row 1) achieving 86.44\% Recall@1. By introducing Self-Distillation (SD), performance improves significantly to 91.46\%. Following this, integrating the UAFA module further advances the Recall@1 to 91.93\%. Finally, Cross-Distillation (CD) elevates performance to 93.87\%. As visualized in Figure~\ref{t-sne} and Figure~\ref{distance}, our MobileGeo model learns a more discriminative feature distribution compared to the baseline.

\textbf{Effectiveness of the Inference-Phase Module.} After establishing a strong student model through our training strategy, we evaluate the contribution of the MSRM. The final row of Table~\ref{ab} shows the impact of applying MSRM to the descriptors generated by our fully trained model. The result is a remarkable jump in performance to 97.15\% Recall@1. This +3.28\% gain over the already powerful base model demonstrates that by refining the feature set at inference time, MSRM significantly enhances localization precision without altering the underlying network architecture. 
% which is applied exclusively during inference. 
% is attributed solely to this lightweight post-processing step. 

\textbf{Edge Device Deployment.} To demonstrate the practical applicability of our approach, as shown in Figure~\ref{onboard}, we evaluate MobileGeo on two representative edge platforms: NVIDIA Jetson TX2 and AGX Orin. As shown in Figure~\ref{tx2}, our method achieves exceptional efficiency with only 4.45 GFLOPs, representing a 95.1\% reduction compared to DAC. This dramatic reduction translates directly to superior real-time performance: MobileGeo achieves 27.1 FPS on the resource-constrained TX2 (7.7$\times$ faster than DAC's 3.5 FPS) and an impressive 251.5 FPS on AGX Orin (6.1$\times$ faster than DAC).

\begin{table}[!t]
\renewcommand{\arraystretch}{1}
\centering
\caption{ABLATION STUDY OF EACH COMPONENT ON THE PERFORMANCE OF Our PROPOSED MobileGeo.}
\label{ab}
\resizebox{0.4\textwidth}{!}{
\begin{tabular}{cccccccc}

\toprule[1.5pt]
\toprule 
\multicolumn{3}{c}{\textbf{HD-CVGL}} & \multirow{2}{*}{\textbf{MSRM}} & \multicolumn{2}{c}{\textbf{Drone$\rightarrow$Satellite}} & \multicolumn{2}{c}{\textbf{Satellite$\rightarrow$Drone}} \\
\cline{1-3}
\cline{5-8}
\textbf{SD} & \textbf{UAFA} & \textbf{CD} & ~  &\textbf{Recall@1} $\uparrow$ &\textbf{AP} $\uparrow$  &\textbf{Recall@1} $\uparrow$ &\textbf{AP} $\uparrow$  \\
\midrule
\ding{55} & \ding{55} & \ding{55} & \ding{55} & 86.44 & 88.69 & 93.72 & 85.13 \\
\ding{51} & \ding{55} & \ding{55} & \ding{55} & 91.46 & 92.91 & 94.57 & 90.49 \\
\ding{51} & \ding{51} & \ding{55} & \ding{55} & 91.93 & 93.29 & 95.14 & 91.35 \\
\ding{51} & \ding{51} & \ding{51} & \ding{55} & \textcolor{blue}{\textbf{93.87}} & \textcolor{blue}{\textbf{94.83}} & \textcolor{red}{\textbf{95.72}} & \textcolor{blue}{\textbf{92.57}} \\
\ding{51} & \ding{51} & \ding{51} & \ding{51} & \textcolor{red}{\textbf{97.15}} & \textcolor{red}{\textbf{97.50}} & \textcolor{blue}{\textbf{95.58}} & \textcolor{red}{\textbf{96.27}} \\
\bottomrule
\bottomrule[1.5pt]

\end{tabular}}

\label{tab:each_component}
\end{table}

% Table 7
% \begin{table}[!t]
% \renewcommand{\arraystretch}{1}
% \centering
% \caption{ABLATION STUDY OF EACH COMPONENT ON THE PERFORMANCE OF Our PROPOSED MobileGeo.}
% \label{edge}
% \resizebox{0.5\textwidth}{!}{
% \begin{tabular}{ccccc}

% \toprule[1.5pt]
% \toprule 
% \multicolumn{1}{c}{\multirow{2}{*}{Method}} & \multicolumn{1}{c}{Params} & \multicolumn{1}{c}{FLOPs} & \multicolumn{2}{c}{Speed (FPS)~$\uparrow$}                               \\
% \multicolumn{1}{c}{}                        & \multicolumn{1}{c}{(M)~$\downarrow$}    & \multicolumn{1}{c}{(G)~$\downarrow$}   & \multicolumn{1}{c}{Jetson TX2} & \multicolumn{1}{c}{AGX Orin} \\

% \midrule
% DAC~\cite{xia2024enhancing}         &   96.50                         &   90.24                        & 3.5                            & 41.0                         \\
% MEAN~\cite{chen2025multi}                                         &  \textcolor{blue}{\textbf{36.50}}                          &  \textcolor{blue}{\textbf{26.18}}                         & \textcolor{blue}{\textbf{9.2}}                            & \textcolor{blue}{\textbf{88.9}}                         \\
% MobileGeo(Ours)                                  & \textcolor{red}{\textbf{28.57}}                           &  \textcolor{red}{\textbf{4.45}}                         & \textcolor{red}{\textbf{27.1}}                           & \textcolor{red}{\textbf{251.5}}   \\
% \bottomrule
% \bottomrule[1.5pt]                    
% \end{tabular}}
% \end{table}

\section{Conclusion}

In this paper, we introduced a mobile-friendly framework MobileGeo, which shifts computational complexity to the training stage, enabling superior performance on resource-constrained devices. We achieve this through two innovations: 1) A comprehensive Hierarchical Distillation (HD-CVGL) strategy during training, which incorporates our Uncertainty-Aware Prediction Alignment (UAPA) to robustly handle data imbalance and domain discrepancies, producing a highly discriminative yet compact student network without any inference overhead. 2) A lightweight Multi-view Selection Refinement Module (MSRM) at inference, which uses mutual information theory to select and fuse the most informative views, boosting accuracy while minimizing feature matching cost. Although this paper focuses on image modalities (drone and satellite imagery), our future work will extend the framework to incorporate additional multimedia inputs, such as infrared images and video data, to better handle real-world extreme scenarios. 

% like nighttime and adverse weather conditions.

% Although this paper focuses on image modalities (drone and satellite imagery), our future research will explore more generalized multimedia information, such as infrared images, to address real-world extreme scenarios like nighttime conditions.

% Experiments show MobileGeo sets a new state-of-the-art, achieving 97.15\% Recall@1 on University-1652 while being over $5\times$ more FLOPs-efficient and $3\times$ faster than previous methods. 

\bibliographystyle{IEEEtran}
\bibliography{mybibfile}

\vfill
\end{document}